\def\year{2022}\relax
\documentclass[letterpaper]{article} 
\usepackage{aaai22}  
\usepackage{times}  
\usepackage{helvet}  
\usepackage{courier}  
\usepackage[hyphens]{url}  
\usepackage{graphicx} 
\usepackage{natbib}  
\usepackage{caption} 
\DeclareCaptionStyle{ruled}{labelfont=normalfont,labelsep=colon,strut=off} 
\usepackage{algorithm}
\usepackage{algorithmic}
\usepackage{amsfonts}
\usepackage{amsmath}
\usepackage{amssymb}
\usepackage{mathtools}
\usepackage{multirow}
\usepackage{times}
\usepackage{epsfig}
\usepackage{graphicx}
\usepackage{subcaption}
\usepackage{gensymb}
\usepackage[super]{nth}
\newcommand{\tabincell}[2]{\begin{tabular}{@{}#1@{}}#2\end{tabular}}

\usepackage{pifont}
\newcommand{\cmark}{\ding{51}}%
\newcommand{\xmark}{\ding{55}}%

\usepackage{color}
\usepackage{colortbl}
\usepackage{comment}

\usepackage{newfloat}
\usepackage{listings}
\floatstyle{ruled}
\newfloat{listing}{tb}{lst}{}
\floatname{listing}{Listing}
\title{AFDetV2: Rethinking the Necessity of the Second Stage for
\\Object Detection from Point Clouds}
\author {
    Yihan Hu\equalcontrib,
    Zhuangzhuang Ding\equalcontrib,
    Runzhou Ge\equalcontrib\\
    Wenxin Shao,
    Li Huang,
    Kun Li,
    Qiang Liu
}
\affiliations {
    Horizon Robotics\\
    \{yihan.hu96, dinghouzx, runzhouge, proliu\}@gmail.com
}

\begin{document}

\maketitle

\newcommand\blfootnote[1]{%
  \begingroup
  \renewcommand\thefootnote{}\footnote{#1}%
  \addtocounter{footnote}{-1}%
  \endgroup
}

\begin{abstract}
There have been two streams in the 3D detection from point clouds: single-stage methods and two-stage methods. While the former is more computationally efficient, the latter usually provides better detection accuracy. By carefully examining the two-stage approaches, we have found that if appropriately designed, the first stage can produce accurate box regression. In this scenario, the second stage mainly rescores the boxes such that the boxes with better localization get selected. From this observation, we have devised a single-stage anchor-free network that can fulfill these requirements. This network, named AFDetV2, extends the previous work by incorporating a self-calibrated convolution block in the backbone, a keypoint auxiliary supervision, and an IoU prediction branch in the multi-task head. We take a simple product of the predicted IoU score with the classification heatmap to form the final classification confidence. The enhanced backbone strengthens the box localization capability, and the rescoring approach effectively joins the object presence confidence and the box regression accuracy. As a result, the detection accuracy is drastically boosted in the single-stage. To evaluate our approach, we have conducted extensive experiments on the Waymo Open Dataset and the nuScenes Dataset. We have observed that our AFDetV2 achieves the state-of-the-art results on these two datasets, superior to all the prior arts, including both the single-stage and the two-stage 3D detectors. AFDetV2 won the \nth{1} place in the Real-Time 3D Detection of the Waymo Open Dataset Challenge 2021. In addition, a variant of our model AFDetV2-Base was entitled the ``Most Efficient Model'' by the Challenge Sponsor, showing a superior computational efficiency. To demonstrate the generality of this single-stage method, we have also applied it to the first stage of the two-stage networks. Without exception, the results show that with the strengthened backbone and the rescoring approach, the second stage refinement is no longer needed.
\end{abstract}


\begin{figure*}[t]
\centering
\includegraphics[width=0.85\textwidth]{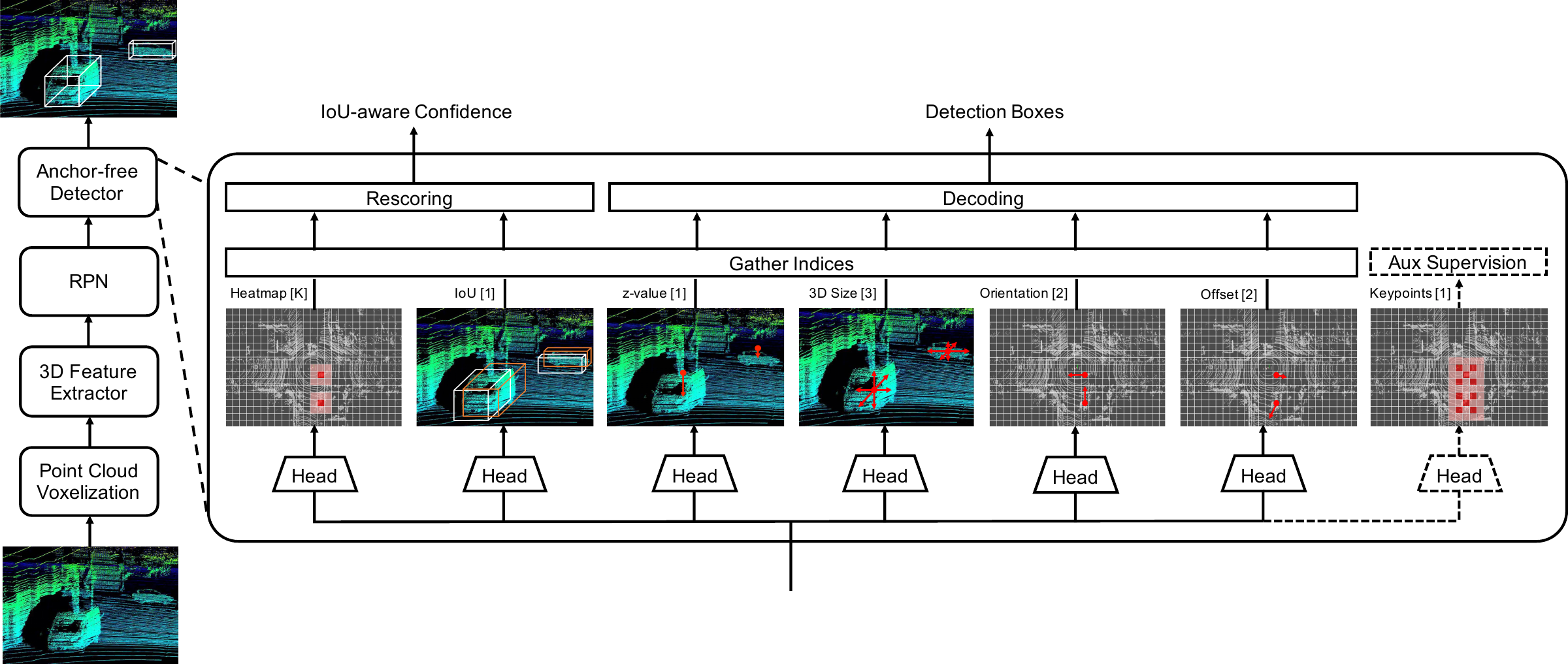} 
\caption{The framework of anchor-free one-stage 3D detection (AFDetV2) system. The whole pipeline consists of the Point Cloud Voxelization, 3D Feature Extractor, backbone and the Anchor-Free Detector. The number in the brackets indicates the output channels in the last convolution layer. $K$ is the number of categories used in the detection. The auxiliary supervision which is turned off at inference is shown in dashed lines. Better viewed in color and zoom in for more details.}
\label{fig:framework}
\end{figure*}

\section{Introduction}

Object detection from point clouds has become a practical solution to robotics vision, especially in autonomous driving applications. Like the detection methods on 2D images, the 3D detection methods can also be divided into two groups: single-stage~\cite{ge2020afdet, zhou2018voxelnet, yan2018second, he2020structure, zheng2020cia, Lang_2019_CVPR, bewley2020range, fan2021rangedet, chen2020every} and two-stage~\cite{shi2019pointrcnn,  yang2019std, qi2017pointnet++, shi2020part, li2021lidar, shi2021pv, deng2021voxel, Yin_2021_CVPR, Sun_2021_CVPR}, in terms of the model structure. The two-stage methods usually show better accuracy~\cite{shi2019pointrcnn, shi2021pv, deng2021voxel} in the classification confidence and the box regression, than the single-stage methods. In these methods, the first stage provides proposals of the bounding boxes, based on which the second stage pools feature and runs through a smaller network to classify it and refine the box regression. The features utilized by the second stage can be extracted from the feature maps (voxel-based) produced by the backbone~\cite{deng2021voxel, Yin_2021_CVPR}, or from the raw point cloud (point-based) which need to go through another round of feature encoding~\cite{yang2019std, shi2019pointrcnn}.

Why is the second stage needed? One argument is that the point features may recover the loss of positioning information, due to voxelization, striding operations, or lack of receptive field. Another argument is that the category classification and the box regression are usually realized in two separate branches, so the confidence map from the classification branch may not align well with the localization accuracy. These arguments are somewhat supported by the evidence that the second stage does boost detection accuracy.

But we wonder whether the above arguments mean the necessity of a second stage. For example, is the raw point necessary for a precise positioning? Some recent two-stage methods~\cite{deng2021voxel, Yin_2021_CVPR} suggest that voxel-based features could achieve the same positioning accuracy without using the point-based features~\cite{shi2019pointrcnn}. What the second stage provides is to refine the classification score and enhance box regression with additional computational blocks. However, after a careful examination, we have found that the first stage is already capable of producing accurate box localization, when designed properly (Sec.~\ref{subsection:two_stage_methods}). The actual contribution from the second stage lies in the enhancement of the classification scores.  In another word, the second stage may not refine the positions of the boxes; instead, it rescores the classification confidence to better select the boxes.

This observation is consistent with the argument of classification-localization misalignment, in that the boxes with better IoU may be associated with a better score at the second stage. But, “can one resolve the misalignment problem within a single-stage approach?”. In fact, some previous works also asked the same question. For example, a recent 2D detection algorithm~\cite{zhang2020varifocalnet} proposes to learn an IoU-aware classification score to join the object presence confidence with the localization accuracy. And, a 3D detection algorithm~\cite{zheng2020cia} proposes an IoU regression branch, the output of which is combined with the classification score for the rescoring.

Our work is closely related to~\cite{zheng2020cia} with a few major differences: first, our single-stage approach is an anchor-free network, extended from the 2020 award-winning work AFDet~\cite{ge2020afdet}, while~\cite{zheng2020cia} is anchor-based; second, we deploy a self-calibrated convolutional block with multi-scalability and spatial attention~\cite{liu2020improving} in our backbone to enlarge the receptive field and improve the semantics, which is simple to plug into an existing network; third, to further enhance the box regression, we employ an auxiliary loss on the 4-corners and the center (\textit{i.e.} keypoints) of the 3D boxes; and fourth, we conduct extensive experiments with both single-stage and two-stage approaches, to reveal that refining the classification confidence and box regression in a second stage is unnecessary. To align the classification score with the box regression, we have also utilized an IoU prediction branch, but with a modified rescoring formula to combine the classification heatmap and the IoU prediction map.

To demonstrate the effectiveness of our single-stage 3D detector, which is named AFDetV2, we have compared with other state-of-the-art models, including both single-stage and two-stage methods, on the Waymo Open Dataset (WOD) and the nuScenes Dataset. On these two datasets, all the experimental results show that AFDetV2 is superior to the prior arts. Our AFDetV2 with the \textit{test} set accuracy of 73.12 APH/L2 and the latency of 60.06 ms won \emph{the \nth{1} place in the Real-Time 3D Detection of the WOD Challenge 2021\textsuperscript{\rm 1}}. \blfootnote{\textsuperscript{\rm 1}\url{http://cvpr2021.wad.vision/}, accessed on Dec 6, 2021.} In addition, a variant of our model AFDetV2-Base with the accuracy of 72.57 APH/L2 and latency of 55.86 ms was entitled \emph{the ``Most Efficient Model'' by the WOD Challenge Sponsor}, showing a superior computational efficiency.

Also, we show that by plugging in the proposed components, \textit{i.e.} self-calibrated block, keypoint auxiliary loss, and the IoU prediction branch, to the first stage of a two-stage approach, such as~\cite{deng2021voxel}, one can discard the second stage and still achieve similar or even better detection accuracy.


This arXiv paper, which has more comparisons, results and details than our original AAAI-22 paper\textsuperscript{\rm 2}\blfootnote{\textsuperscript{\rm 2}\url{https://ojs.aaai.org/index.php/AAAI/article/view/19980}, accessed on Jul 17, 2022.} , is the complete version of this paper.



\section{Related Work}
\subsection{Two-Stage/Singe-Stage LiDAR Detector}
Inspired by 2D detection~\cite{girshick2015fast, ren2016faster, cai2018cascade}, a two-stage LiDAR detector usually generates Region of Interests (RoIs) at the first stage, followed by a second stage to refine the first stage predictions. PointRCNN~\cite{shi2019pointrcnn} and STD~\cite{yang2019std} apply R-CNN style detector from 2D to 3D domain. After generating coarse 3D bounding box proposals using PointNet++~\cite{qi2017pointnet++}, point features within 3D proposals are directly pooled to second stage for refinement. However, different proposals might end up pooling the same group of points, which loses the ability to encode the geometric information of the proposals. To tackle this problem, Part-$A^2$~\cite{shi2020part} designed an RoI-aware point cloud pooling operation, while LiDAR R-CNN~\cite{li2021lidar} devised a method with  virtual point and boundary offset. Another point pooling method proposed by PV-RCNN~\cite{shi2021pv} summarizes learned point and voxel-wise feature volumes at multiple neural layers into a small set of keypoints, then the keypoint features are aggregated according to RoI-grid. Pyramid R-CNN~\cite{mao2021pyramid} utilizes an RoI-grid pyramid to mitigate the sparsity problem. Instead of pooling from point features, Voxel R-CNN\cite{deng2021voxel} designs voxel-RoI pooling module to directly pool from voxel and BEV feature space according to the RoI-grid. To speed up, CenterPoint~\cite{Yin_2021_CVPR} simplifies the pooling module by sampling five keypoints from BEV features using bilinear interpolation. Recently, RSN~\cite{Sun_2021_CVPR} utilizes a foreground segmentation as a first stage to sparsify the point clouds, which boosts the efficiency of the second stage sparse convolution.

For single-stage LiDAR detectors, VoxelNet~\cite{zhou2018voxelnet} encodes the point cloud data as 3D voxels and uses 3D convolution to extract 3D features. However, 3D convolution is computationally expensive. Considering the sparsity nature of the point clouds, SECOND~\cite{yan2018second} utilizes 3D sparse convolution to speed up the 3D convolution. To speed up the encoding process, PointPillars~\cite{Lang_2019_CVPR} encodes 3D point cloud data as BEV pillars, then conventional 2D CNN can be applied to the pseudo image. CIA-SSD~\cite{zheng2020cia} utilizes an IoU prediction branch and a post-processing method to incorporate localization accuracy to confident scores.  AFDet~\cite{ge2020afdet, ge2021real, ding20201st, wang20201st}, which served as the base detector for several \nth{1} place winner solutions of Waymo Open Dataset Challenge 2020, proposes an anchor-free and NMS free LiDAR detection framework for the first time.  Recently, some works focus on the representation of the different views. \citeauthor{bewley2020range, fan2021rangedet} exploit LiDAR detection in the range view. \citeauthor{chen2020every, zhou_2020_mvf} fuse the complementary information from the range view and the BEV.

\subsection{Anchor-Free/Anchor-Based LiDAR Detector}
The idea of anchor first appears in Faster R-CNN~\cite{ren2016faster}. Inspired by Ren, multiple works~\cite{Lang_2019_CVPR, yan2018second, zhou2018voxelnet, kuang2020voxel, zhu2019class, qi2019deep, zhou_2020_mvf, chen2017multi, noh2021hvpr} utilize anchors to improve the performance of their LiDAR 3D detection networks. Different from 2D networks, anchors are extended into 3D space with a z-axis value. \citeauthor{yang2019std} proposes a new spherical anchor to seed each point more efficiently and propose objects with higher recall. \citeauthor{shi2020part} investigates both anchor-free and anchor-based strategies, in which anchor-based strategy achieves higher recall rates sacrificing memory and calculation costs. To improve the efficiency of networks, some works remove the process of anchor assignment. PointRCNN~\cite{shi2019pointrcnn} uses a foreground segmentation network to propose objects which can significantly reduce the number of proposals. PIXOR~\cite{yang2018pixor} assigns all pixels inside ground truth bounding boxes as positive samples in the BEV feature map. Similarly, some range-view-based methods assign pixels on range view maps that are inside 3D ground truth boxes as positive samples~\cite{fan2021rangedet, meyer2019lasernet, bewley2020range}. Multiple anchor-free detection networks formulate the detection problem as a keypoint detection problem~\cite{ge2020afdet, Sun_2021_CVPR, Yin_2021_challenge}. All features are acquired according to the ``centerness'' of objects so anchors can be deprecated. 3D-MAN~\cite{yang20213d} adopted an anchor-free single-stage detector as the first module of its multi-frame architecture to generate high-quality proposals.

Compared to the two-stage and anchor-based 3D detectors, the single-stage and anchor-free 3D detector has a much simpler structure and higher speed, which is more suitable for real-time deployment. Here we propose AFDetV2, a fast and accurate anchor-free single-stage 3D detector, which surpasses all two-stage detectors in terms of accuracy and speed.

\begin{figure}[t]
\hfill
\begin{subfigure}[b]{.4\columnwidth}
\centering
\includegraphics[width=\linewidth]{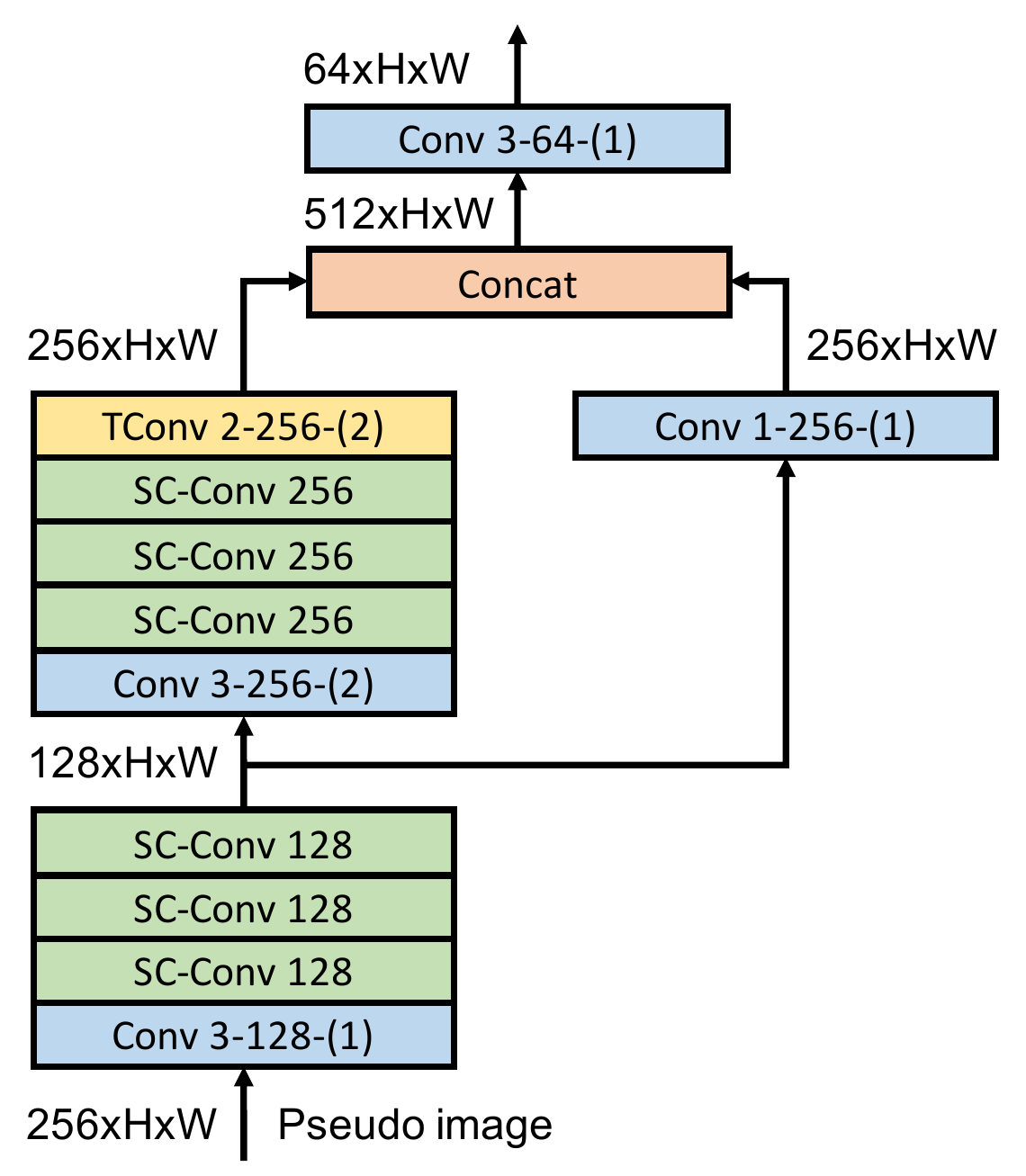}
\caption{SC-Conv backbone}\label{fig:RPN2}
\end{subfigure}
\hfill
\begin{subfigure}[b]{.56\columnwidth}
\centering
\includegraphics[width=\linewidth]{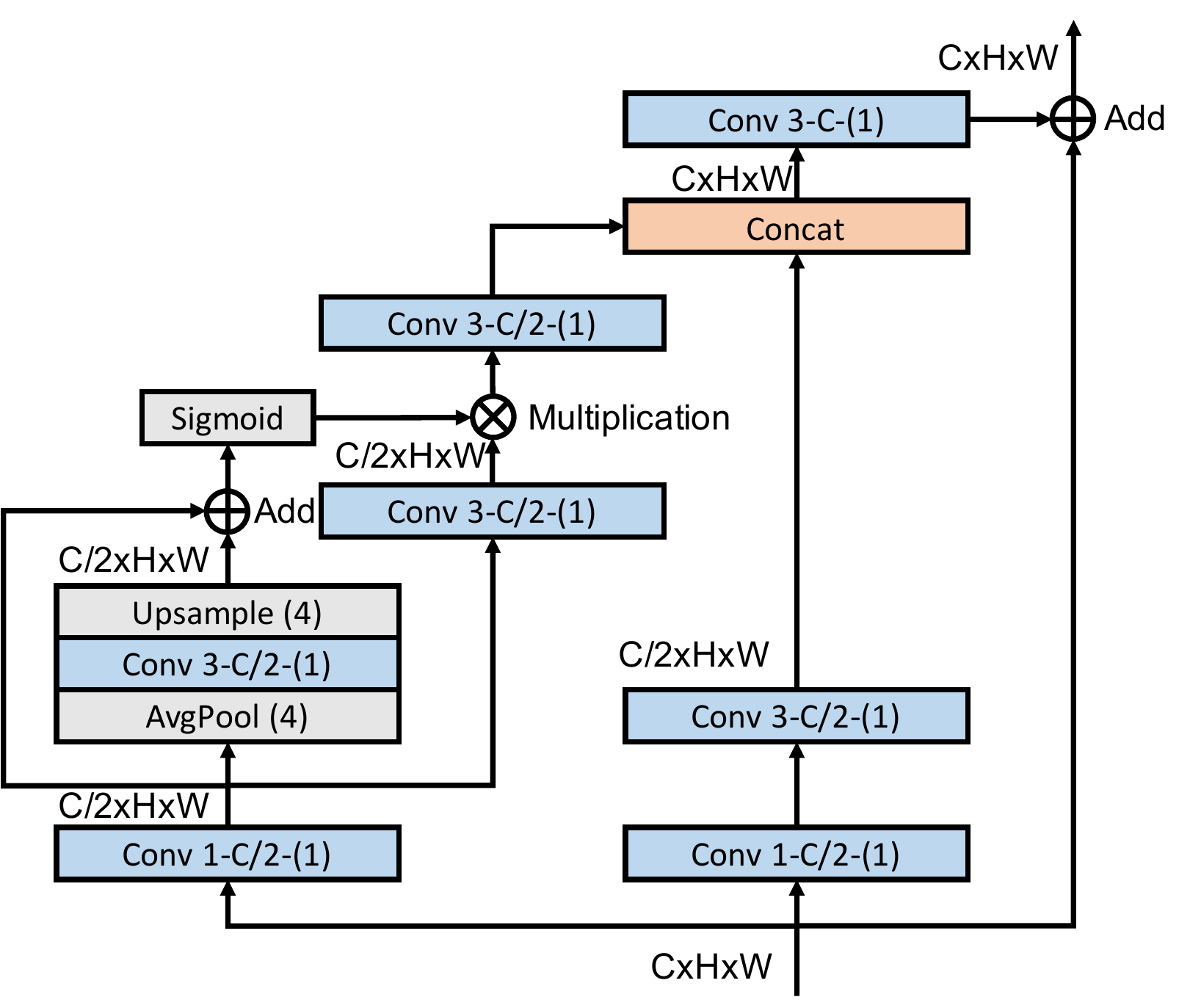}
\caption{SC-Conv module}\label{fig:SCConv}
\end{subfigure}
\caption{(a) denotes the backbone with the self-calibrated convolution (SC-Conv). ``Conv" stands for convolutional layer and ``TConv" stands for transposed convolutional layer. The format of the layer setting follows ``kernel size-channels-(strides)", \textit{i.e.} $k$-$C$-($s$). ``SC-Conv" stands for SC-Conv module and only channels are shown. (b) gives the detailed structure of the SC-Conv module. }
\label{fig:RPN}
\end{figure}


\section{Methods}
 In this section, we first analyze the two-stage methods, focusing on the necessity of the second stage. Then we present our single-stage method.


\begin{table*}[h]
\begin{center}
\resizebox{0.75\textwidth}{!}{
\vspace{5pt}
\setlength\tabcolsep{2pt}
\begin{tabular}{l|c|c|c c c|c c c|c c c|c c c}
\hline
\multirow{2}{*}{Variants} &\multirow{2}{*}{Sensors} &\multirow{2}{*}{Frames} &\multirow{2}{*}{Ens} & Usage  &Server &\multicolumn{3}{c|}{Training Range} &\multicolumn{3}{c|}{Inference Range}  &\multicolumn{3}{c}{Grid Size} \\
&&&&Scenario &Latency &$x$ &$y$ &$z$ &$x$ &$y$ &$z$ &$x$ &$y$ &$z$\\
\hline
\hline
AFDetV2-Lite  & LT &1  &\xmark &Onboard  &46.90     &$\pm$75.2  &$\pm$75.2  &(-2, 4)  &$\pm$75.2  &$\pm$75.2  &(-2, 4) &0.1  &0.1  &0.15  \\
AFDetV2-Base  & LT &2  &\xmark &Onboard  &55.86     &$\pm$75.2  &$\pm$75.2  &(-2, 4)  &$\pm$75.2  &$\pm$75.2  &(-2, 4) &0.1  &0.1  &0.15  \\
AFDetV2       & LT &2  &\xmark &Onboard  &60.06     &$\pm$75.2  &$\pm$73.6  &(-2, 4)  &$\pm$80.0  &$\pm$76.16 &(-2, 4) &0.1  &0.08 &0.15  \\
AFDetV2-Ens   & L  &2  &\cmark &Offboard &-         &$\pm$75.2  &$\pm$73.6  &(-2, 4)  &$\pm$80.0  &$\pm$76.16 &(-2, 4) &0.1  &0.08 &0.15  \\

\hline
\end{tabular}
}
\end{center}
\caption{The configurations of different variants of our model for Waymo Open Dataset. ``L" and ``LT" mean ``all LiDARs" and ``top-LiDAR only", separately. ``Ens" is short for ensemble. ``Server Latency" is the latency measured by the official testing server in milliseconds. Values in training range, inference range and grid size columns are in meters with respect to $x, y, z$-axes, respectively. Our AFDetV2-Ens uses Test-Time Augmentation and model ensemble for better detection accuracy and is suitable for offboard usage (\textit{e.g.} auto labeling). The other 3 variants are non-ensemble, real-time and suitable for onboard usage.}
\label{tbl:model_training}
\end{table*}
\begin{table*}[t]
    \definecolor{Gray}{gray}{0.9}
    \newcolumntype{g}{>{\columncolor{Gray}}c}
    \small
    \begin{center}
		\resizebox{\textwidth}{!}{
            \begin{tabular}{c|c|c|gc|cccccccccc}
                \hline
            
                \tabincell{c}{Methods}
                & Reference & \#Stages & NDS & mAP & Car & Truck & Bus & Trailer & Cons. Veh. & Ped. & Motor. & Bicycle & Tr. Cone & Barrier    \\
                \hline
                \hline
                WYSIWYG & CVPR~\citeyear{hu2020you} & Single & 41.9 & 35.0 & 79.1 & 30.4 & 46.6 & 40.1 & 7.1 & 65.0 & 18.2 & 0.1 & 28.8 & 34.7\\                
                PointPillars & CVPR~\citeyear{Lang_2019_CVPR} & Single & 45.3 & 30.5 & 68.4 & 23.0 & 28.2 & 23.4 & 4.1 & 59.7 & 27.4 & 1.1 & 30.8 & 38.9\\
                3DVID & CVPR~\citeyear{yin2020lidar} & Single & 53.1 & 45.4 & 79.7 & 33.6 & 47.1 & 43.1 & 18.1 & 76.5 & 40.7 & 7.9 & 58.8 & 48.8 \\
                3DSSD & CVPR~\citeyear{yang20203dssd} & Single & 56.4 & 42.6 & 81.2 & 47.2 & 61.4 & 30.5 & 12.6 & 70.2 & 36.0 & 8.6 & 31.1 & 47.9  \\
                Cylinder3D & TPAMI~\citeyear{zhu2021cylindrical}       & Single & 61.6 & 50.6 & - & - & - & - & - & - &- & - & - & - \\
                CGBS & arXiv~\citeyear{zhu2019class} & Single & 63.3 & 52.8& 81.1& 48.5 &54.9 &42.9 &10.5& 80.1 & 51.5& 22.3& 70.9 & 65.7 \\
                Pointformer & CVPR~\citeyear{pan20213d} & Single & - & 53.6 & 82.3 & 48.1 & 55.6 & 43.4 & 8.6 & 81.8 & 55.0 & 22.7 & 72.2 & 66.0 \\
                CVCNet & NeurIPS~\citeyear{chen2020every} & Single  & 64.2 & 55.8 & 82.6 & 49.5 & 59.4 & 51.1 & 16.2 & 83.0 & 61.8 & 38.8 & 69.7 & 69.7 \\  
                CenterPoint & CVPR~\citeyear{Yin_2021_CVPR} & Two & 65.5 & 58.0 & 84.6 & 51.0 & 60.2 & 53.2 & 17.5 & 83.4 & 53.7 & 28.7 & 76.7 & 70.9\\
                HotSpotNet & ECCV~\citeyear{chen2020object} & Single & 66.0 & 59.3 & 83.1 & 50.9 & 56.4 & 53.3 & 23.0 & 81.3 & 63.5 & \bf36.6 & 73.0 & \bf71.6\\
                AFDetV2 & Ours & Single & \bf68.5 & \bf62.4 & \bf86.3 & \bf54.2 & \bf62.5 & \bf58.9 & \bf26.7 & \bf85.8 & \bf63.8 & 34.3 & \bf80.1 & 71.0 \\
                \hline
            \end{tabular}
        }
    \end{center}
    \caption{The LiDAR-only non-ensemble performance comparison on the  nuScenes \textit{test} set. The table is mainly sorted by nuScenes detection score (NDS) which is the official ranking metric.
    }
    \label{tbl: nusc ablation}
\end{table*}

\subsection{Analysis on Two-stage Methods}
\label{subsection:two_stage_methods}
There are two main causes for a second stage: (1) the RoI proposal from the first stage confines the spatial range so the PointNet based feature extraction can be afforded, which may restore the 3D context in better resolutions; (2) the second stage, as an extra computational block dedicated to an RoI, can refine the classification score and the box regression. 

There are studies countering the former argument. In~\cite{deng2021voxel}, the authors compare the point-based methods, \textit{e.g.}~\cite{shi2019pointrcnn, yang2019std, shi_2020_challenge}, to the voxel-based methods, \textit{e.g.}~\cite{shi2020part, deng2021voxel}, and show that with small voxel size (0.1m, 0.1m, 0.15m), the voxel-based methods can achieve similar accuracy as the point-based. Also, a more recent method~\cite{Yin_2021_CVPR, Yin_2021_challenge} demonstrates the state-of-the-art performances on multiple renowned datasets by using voxel-based features. These all suggest that the second stage may not necessarily provide a better spatial resolution than the first stage.

For the latter, there may be three factors of the possible improvements: the box regression, the classification score, and the alignment between these two. To differentiate the contributions of these factors, we conduct an experiment: we add the classification and the box regression branches to a bare single-stage network in turn and examined the corresponding improvements of the final average precision (AP). We find that adding the box regression alone in the second stage do not show any significance. But adding the score refinement in the second stage does improve the final AP significantly (Tab.~\ref{tbl: ablation 2stage vs 1stage}). This suggests that the second stage may not necessarily enhance the box regression. Instead, it can enhance the classification confidence such that the box with better regression gets a higher score, and vice versa.

So the second stage can enhance the classification score, by properly pulling the context of the features and applying an extra computational block. We wonder if, by properly strengthening the feature extraction of the first stage, we may achieve a similar enhancement. With this thought, we conduct another experiment: we replace the convolutional block of the backbone in~\cite{deng2021voxel} with a self-calibrated convolution block~\cite{liu2020improving}, to strengthen the semantics of the features. We observe that this simple replacement greatly improve the overall AP (Tab.~\ref{tbl: rethinking Voxel R-CNN}). Although still slightly lower than the improvement of using the second stage, this result motivates a further experiment: adding an IoU alignment to the classification. So on top of the backbone enhancement, we add an IoU prediction branch to the first stage. And we combine the IoU prediction score with the classification score by a simple product (Eq.~\ref{eq:iou_recore}). This experiment wins the arguments against the second stage.

With the above observations, we believe a single-stage 3D detector can be as accurate as the current state-of-the-art two-stage 3D detectors. We present our single-stage framework in details in the next few sections.

\begin{table*}
    \definecolor{Gray}{gray}{0.9}
    \newcolumntype{g}{>{\columncolor{Gray}}c}
	\begin{center}
		\resizebox{\textwidth}{!}{
			\begin{tabular}{c|c|c|cg|cc|cc|cc}
				\hline
				\multirow{2}{*}{Methods} & 
				\multirow{2}{*}{Reference} &
				\multirow{2}{*}{\# Stages} & 
				\multicolumn{2}{c|}{ALL (3D AP/APH)} & \multicolumn{2}{c|}{VEH (3D AP/APH)} & 
				\multicolumn{2}{c|}{PED (3D AP/APH)} & \multicolumn{2}{c}{CYC (3D AP/APH)} \\
				&&& L1 & L2 & L1 & L2 & L1 & L2 & L1 & L2 \\
				\hline
				\hline
				StarNet                 & arXiv~\citeyear{ngiam2019starnet}      & Two   & - & - & 53.70/- & - & 66.80/- & - & - & -  \\ 
				PPBA                    & ECCV~\citeyear{cheng2020improving}  & Single & - & - &62.40/- & - & 66.00/- & - & -  & -\\
				MVF                     & CoRL~\citeyear{zhou_2020_mvf}  & Single & - & - & 62.93/- & - & 65.33/- & - & - & -\\
				AFDet                   & CVPRW~\citeyear{ge2020afdet}   & Single & - & - & 63.69/- & - & - & - & - & - \\
				CVCNet                  & NeurIPS~\citeyear{chen2020every}      & Single & - & - & 65.20/- & - & - & - & - & - \\
				Perspective EdgeConv            & CVPR~\citeyear{chai2021point}     & Single & - & - & 65.20/- & -/56.70 & 73.90/- & -/59.60 & - & - \\
				3D-MAN                  & CVPR~\citeyear{yang20213d}     & Multi      & - & - & 69.03/68.52 & 60.16/59.71 & 71.71/67.74 & 62.58/59.04 & - & - \\
				RCD                     & CoRL~\citeyear{bewley2020range}      & Two    & - & - & 69.59/69.16 & - & - & - & - & - \\
				Pillar-based            & ECCV~\citeyear{wang2020pillar}      & Single & - & - & 69.80/- & - & 72.51/- & - & - & - \\
				$^\dagger$SECOND        & Sensors~\citeyear{yan2018second}      & Single      & 67.20/63.05 & 60.97/57.23 & 72.27/71.69 & 63.85/63.33 & 68.70/58.18 & 60.72/51.31 & 60.62/59.28 & 58.34/57.05 \\
				$^\ddagger$PointPillars & CVPR~\citeyear{Lang_2019_CVPR}  & Single      &  68.87/63.33 & 62.63/57.53 & 71.60/71.00 & 63.10/62.50 & 70.60/56.70 & 62.90/50.20 & 64.40/62.30 & 61.90/59.90 \\
				LiDAR R-CNN             & CVPR~\citeyear{li2021lidar}      & Two    & 71.10/66.20 & 64.63/60.10  & 73.50/73.00 & 64.70/64.20 & 71.20/58.70 & 63.10/51.70 & 68.60/66.90 & 66.10/64.40 \\
				RangeDet                & arXiv~\citeyear{fan2021rangedet}      & Single & 71.53/- & - &72.85/ & - & 75.94/- & - & 65.80/- & - \\
				MVF++                   & CVPR~\citeyear{Qi_2021_CVPR}      & Single & - & - & 74.64/- & - & 78.01/- & - & - & - \\
				Voxel Transformer       & ICCV~\citeyear{mao2021voxel}      & Two    & - & - &74.95/74.25 & 65.91/65.29 & - & - & - & - \\
				RSN                     & CVPR~\citeyear{Sun_2021_CVPR}      & Two    & - & - & 75.10/74.60 & 66.00/65.50 & 77.80/72.70 & 68.30/63.70 & - & - \\
				$\text{H}^2$3D R-CNN    & TCSVT~\citeyear{deng2021multi}    & Two & - & - & 75.15/- & 66.14/- & - & - & - & - \\
				M3DETR                  & arXiv~\citeyear{guan2021m3detr}        & Two    & - & - & 75.71/75.08 & 66.58/66.02 & - & - & - & - \\
				Voxel R-CNN             & AAAI~\citeyear{deng2021voxel}         & Two    & - & - & 75.59/- & 66.59/- & - & - & - & - \\
				Pyramid R-CNN           & ICCV~\citeyear{mao2021pyramid}        & Two    & - & - & 76.30/75.68 & 67.23/66.68 & - & - & - & - \\
				CenterPoint             & CVPR~\citeyear{Yin_2021_CVPR}        & Two    & - & - & 76.70/76.20 & 68.80/68.30 & 79.00/72.90 & 71.00/65.30 & - & - \\
				PV-RCNN                 & CVPR~\citeyear{shi2020pv}         & Two    & 73.44/69.63 & 66.80/63.33 & 77.51/76.89 & 68.98/68.41 & 75.01/65.65 & 66.04/57.61 & 67.81/66.35 & 65.39/63.98 \\
				Part-$A^2$              & TPAMI~\citeyear{shi2020part}          & Two     & 73.63/70.25 & 66.93/63.84 & 77.05/76.51 & 68.47/67.97 & 75.24/66.87 & 66.18/58.62 & 68.60/67.36 & 66.13/64.93 \\
				CT3D                    & ICCV~\citeyear{sheng2021improving}          & Two & - & - & - & 69.04/- & - & - & - & - \\
				PV-RCNN-v2              & arXiv~\citeyear{shi2021pv}          & Two    & 74.81/71.00 & 68.42/64.87 & \textbf{78.79/78.21} & \textbf{70.26/69.71} & 76.67/67.15 & 68.51/59.72 & 68.98/67.63 & 66.48/65.17 \\
				AFDetV2-Lite            & Ours    & Single & \textbf{77.18/74.83} & \textbf{70.97/68.77} & 77.64/77.14 & 69.68/69.22 & \textbf{80.19/74.62} & \textbf{72.16/66.95} & \textbf{73.72/72.74} & \textbf{71.06/70.12} \\
				\hline
			\end{tabular}
		}
	\end{center}
	\caption{The single-frame LiDAR-only non-ensemble 3D AP/APH performance comparison on the Waymo Open Dataset \textit{val} set. ``ALL" stands for the mean of all three classes. The table is mainly sorted by ALL APH/L2 which is the official ranking metric. $\dagger$: reported by PV-RCNN++~\shortcite{shi2021pv}. $\ddagger$: reported by LiDAR R-CNN~\shortcite{li2021lidar}.}
	\label{tab:wod_val_3d}
\end{table*} 

\subsection{Backbone} 
Our overall network structure follows the previous work AFDet~\cite{ge2020afdet}, in which an anchor-free single-stage 3D detector is proposed. The whole network takes a point cloud input and voxelizes it. The voxel features are then sent into a 3D feature extractor~\cite{ge2021real}, followed by the backbone, and finally into an anchor-free detection head. In this section, we present the details of the first three steps. We present anchor-free detection head in Sec.~\ref{subsection:af_head}.

\subsubsection{Encoder}
We first voxelize~\cite{zhou2018voxelnet, zhu2019class} points into small voxels across 3D space. For this step, the position of voxels is determined by pre-defined grid size. In each voxel, the mean of all points is calculated and is used as the representative value. Thus all points with coordinate are quantized into fixed voxels.

After voxelization, inputs are sent into a 3D Feature Extractor. The 3D Feature Extractor is composed of 3D sparse convolutional layers~\cite{yan2018second, ge2021real} and sub-manifold sparse convolutional layers~\cite{graham2017submanifold}. The extractor is designed to have fewer residual blocks with slightly more channels at the early stage. We set the stride of $z$-axis dimension to 8 to be more efficient. At the end of the 3D Feature Extractor, the resulting feature map is reshaped to form a BEV pseudo image.

\subsubsection{Self-calibrated convolutional backbone}
After 3D feature extraction, the feature map is sent into a multi-scale backbone. To fully explore the potential of the single-stage framework, we apply the Self-Calibrated Convolutions (SC-Conv) \cite{liu2020improving} to the backbone and replace the basic $3 \times 3$ convolutional blocks. SC-Conv block efficiently enlarges the receptive field and adds channel-wise and spatial-wise attention, which increases the detection accuracy without sacrificing the computational costs. The structure of the backbone and SC-Conv block are shown in Fig.~\ref{fig:RPN2} and~\ref{fig:SCConv}.

\begin{table*}
    \definecolor{Gray}{gray}{0.9}
    \newcolumntype{g}{>{\columncolor{Gray}}c}
	\begin{center}
		\resizebox{\textwidth}{!}{
			\begin{tabular}{c|c|c|c|c|cg|cc|cc|cc}
				\hline
				\multirow{2}{*}{Methods} & 
				\multirow{2}{*}{Sensors} &
				\multirow{2}{*}{Frames} &
				\multirow{2}{*}{Ens} &
				Server & 			
				\multicolumn{2}{c|}{ALL (3D AP/APH)} & \multicolumn{2}{c|}{VEH (3D AP/APH)} & 
				\multicolumn{2}{c|}{PED (3D AP/APH)} & \multicolumn{2}{c}{CYC (3D AP/APH)} \\
				&&&& Latency & L1 & L2 & L1 & L2 & L1 & L2 & L1 & L2 \\
				\hline
				\hline
				\multicolumn{13}{c}{(a) Single-frame LiDAR-only Non-ensemble Methods} \\
				StarNet~\shortcite{ngiam2019starnet} & - & 1 & \xmark & - & - & - & 61.50/61.00 & 54.90/54.50 & 67.80/59.90 & 61.10/54.00 & - & -  \\ 
				PPBA~\shortcite{cheng2020improving} & - & 1 & \xmark & - & - & - & 67.50/67.00 & 59.60/59.10 & 69.70/61.70 & 63.00/55.80 & - & - \\
				$^\dagger$PointPillars~\shortcite{Lang_2019_CVPR} & LT & 1 & \xmark & - & - & - & 68.60/68.10 & 60.50/60.10 & 68.00/55.50 & 61.40/50.10 & - & - \\ 
				RCD~\shortcite{bewley2020range} & - & 1 & \xmark & - & - & - & 71.97/71.59 & 65.06/64.70 & - & - & - & - \\
				Pseudo-Labeling~\shortcite{caine2021pseudo}   & - & 1 & \xmark &- & - & - & 74.00/73.60 & - & 69.80/57.90 & - & - & - \\
				M3DETR~\shortcite{guan2021m3detr} & - & 1 & \xmark & - & 71.05/67.09 & 65.50/61.92 & 77.75/77.17 & 70.63/70.06 & 68.10/58.87 & 60.57/52.37 & 67.28/65.69 & 65.31/63.75 \\
				Light-FMFNet~\shortcite{Murhij_2021_challenge} & L & 1 & \xmark & 62.31 & 71.24/67.26 & 65.88/62.18 & 77.85/77.30 & 70.16/69.65 & 69.52/59.78 & 63.62/54.61 & 66.34/64.69 & 63.87/62.28 \\
				HIKVISION\_LiDAR~\shortcite{xu2021centeratt} & L & 1 & \xmark & 54.13 & 75.19/72.58 & 69.74/67.29 & 78.63/78.14 & 71.06/70.60 & 76.00/69.90 & 69.82/64.11 & 70.94/69.70 & 68.35/67.15 \\
				CenterPoint~\shortcite{Yin_2021_CVPR} & - & 1 & \xmark & - & - & - & 80.20/79.70 & 72.20/71.80 & 78.30/72.10 & 72.20/66.40 & - & - \\
				AFDetV2-Lite~(Ours) & LT & 1 & \xmark & \textbf{46.90} & \textbf{77.56/75.20} & \textbf{72.18/69.95} & \textbf{80.49/80.03} & \textbf{72.98/72.55} & \textbf{79.76/74.35} & \textbf{73.71/68.61} & \textbf{72.43/71.23} & \textbf{69.84/68.67} \\
				\hline
				\multicolumn{13}{c}{} \\
				\multicolumn{13}{c}{(b) Multi-frame LiDAR-only Non-ensemble Methods} \\
				3D-MAN~\shortcite{yang20213d} & L & 16 & \xmark & - & - & - & 78.71/78.28 & 70.37/69.98 & 69.97/65.98 & 63.98/60.26 & - & - \\
				RSN~\shortcite{Sun_2021_CVPR} & LT & 3 & \xmark & - & - & - & 80.70/80.30 & 71.90/71.60 & 78.90/75.60 & 70.70/67.80 & - & - \\ 
				X\_Autonomous3D~\shortcite{Liu_2021_challenge} & L & 2 & \xmark & 68.42 & 77.54/75.61 & 72.29/70.46 & 81.49/81.02 & 74.04/73.60 & 78.17/73.93 & 72.29/68.27 & 72.96/71.88 & 70.55/69.50 \\
				CenterPoint~\shortcite{Yin_2021_CVPR} & L & 2 & \xmark & - & 78.71/77.18 & 73.38/71.93 & 81.05/80.59 & 73.42/72.99 & 80.47/77.28 & 74.56/71.52 & 74.60/73.68 & 72.17/71.28 \\
				AFDetV2-Base~(Ours) & LT & 2 & \xmark & \textbf{55.86} & 79.24/77.67 & 74.06/72.57 & 81.27/80.82 & 73.89/73.46 & 81.08/77.87 & 75.34/72.29 & 75.35/74.33 & 72.96/71.97 \\
				Pyramid R-CNN~\shortcite{mao2021pyramid} & L & 2 & \xmark & - & - & - & 81.77/81.32 & 74.87/74.43 & - & - & - & - \\ 
				CenterPoint++~\shortcite{Yin_2021_challenge} & LT & 3 & \xmark & 57.12 & 79.41/77.96 & 74.22/72.82 & \textbf{82.78/82.33} & \textbf{75.47/75.05} & 81.07/\textbf{78.21} & 75.13/72.41 & 74.40/73.33 & 72.04/71.01 \\
				AFDetV2~(Ours) & LT & 2 & \xmark & 60.06 & \textbf{79.77/78.21} & \textbf{74.60/73.12} & 81.65/81.22 & 74.30/73.89 & \textbf{81.26}/78.05 & \textbf{75.47/72.41} & \textbf{76.41/75.37} & \textbf{74.05/73.04} \\
				\hline
				\multicolumn{13}{c}{} \\
				\multicolumn{13}{c}{(c) Ensemble Methods} \\
				TS-LidarDet~\shortcite{Wang_2020_challenge} & L & 1 & \cmark & - & 74.87/71.05 & 69.10/65.53 & 80.75/80.18 & 72.65/72.12 & 74.45/65.01 & 68.10/59.32 & 69.42/67.97 & 66.55/65.16 \\
				RSN Ens~\shortcite{Sun_2021_CVPR} & LT & 3 & \cmark & - & - & - & 81.38/80.97 & 72.80/72.43 & 82.41/77.98 & 74.75/70.68 & - & - \\
				PV-RCNN Ens~\shortcite{shi_2020_challenge} & L & 2 & \cmark & - & 78.82/76.90 & 73.35/71.52 & 81.06/80.57 & 73.69/73.23 & 80.31/76.28 & 73.98/70.16 & 75.10/73.84 & 72.38/71.16 \\
				3DAL~\shortcite{Qi_2021_CVPR} & L & $\sim${200}  & \cmark & - & - & - & \textbf{85.84/85.46} & 77.24/76.91 & - & - & - & - \\ 
				HorizonLiDAR3D~\shortcite{ding20201st} & CL & 5 & \cmark & - & 83.28/81.85 & 78.49/77.11 & 85.09/84.68 & 78.23/77.83 & 85.03/82.10 & 79.32/76.50 & 79.73/78.78 & 77.91/76.98 \\
				AFDetV2-Ens~(Ours) & L & 2 & \cmark & - & \textbf{84.07/82.63} & \textbf{79.04/77.64}  & 85.80/85.41  & \textbf{78.71/78.34} & \textbf{85.22/82.16} & \textbf{79.71/76.75} & \textbf{81.20/80.30} & \textbf{78.70/77.83} \\
				\hline
			\end{tabular}
		}
	\end{center}
	\caption{The 3D AP/APH performance comparison on the Waymo Open Dataset \textit{test} set. ``L", ``LT" and ``CL" mean ``all LiDARs", ``top-LiDAR only" and ``camera and all LiDARs", separately. ``ALL" stands for the mean of all three classes. The table is mainly sorted by ALL APH/L2 which is the official ranking metric. ``Ens" is short for ensemble.  ``Server Latency" is the latency measured by the official testing server in milliseconds. The latency optimization is allowed by the Challenge Sponsor. The table is split into 3 sub-tables: (a) single-frame LiDAR-only non-ensemble methods; (b) multi-frame LiDAR-only non-ensemble methods; (c) ensemble methods. Our models consistently outperform previous state-of-the-art methods under different settings. AFDetV2 and AFDetV2-Base are the two award-winning entries. $\dagger$: reported by RSN~\shortcite{Sun_2021_CVPR}.}
	\label{tab:wod_test_3d}
\end{table*} 

\subsection{Anchor-Free Head}
\label{subsection:af_head}

In addition to the five sub-heads introduced in AFDet~\cite{ge2020afdet}, we devise an IoU-aware confidence score prediction, which is a key to removing the second stage. We also employ a keypoint auxiliary loss to add additional supervision, as shown in Fig.~\ref{fig:framework}. The 5 common sub-heads in both AFDet and AFDetV2 are the heatmap prediction head, the location offset regression head, the $z$-axis location regression head, the 3D object size regression head, and the orientation regression head. Following~\cite{Yin_2021_CVPR}, the minimum allowed Gaussian radius~\cite{law2018cornernet} for the heatmap and keypoint head is set to 2. The regression target is set to $sin$ and $cos$ values of the object yaw angle for the orientation regression sub-head.

\subsubsection{IoU-aware confidence score prediction} The classification score is commonly used as the final prediction score in object detection tasks. However, the classification score lacks the localization information, which is not a good confidence estimate for object detection. Specifically, boxes with high localization accuracy but low classification scores may be deleted after Non-Maximum Suppression. Also, the misalignment harms the ranking-based metrics such as Average Precision. To alleviate the misalignment, most of the existing methods adopt an IoU-aware prediction branch in the second stage network~\cite{jiang2018acquisition,shi2020pv,Yin_2021_CVPR}. However, an additional stage will increase the computational cost and the latency of the network. Also, special operators such as RoI Align~\cite{he2017mask, Yin_2021_CVPR} or RoI pooling~\cite{shi2019pointrcnn,shi2020pv,shi2020part} are required in the second stage network.

Recently, CIA-SSD~\cite{zheng2020cia} migrates IoU-prediction head from 2D image~\cite{wu2020iou} to anchor-based 3D LiDAR detection. Different from CIA-SSD, we adopt an IoU prediction head to our anchor-free network. To incorporate IoU information into confidence score, we recalculate the final confidence score by a simple post-processing function:
\begin{equation}
f = score^{1 - \alpha} * iou ^ {\alpha}
\label{eq:iou_recore}
\end{equation}
where $score$ is the original classification score, $iou$ is the predicted IoU and $\alpha$ is the hyperparameter $\in[0, 1]$ that controls the contributions from the classification score and predicted IoU. 

After rescoring, the ranking of the predictions takes both the classification confidence and localization accuracy into account. The rescoring process will lower the confidence of the predictions with higher classification scores but worse localization accuracy, and vice versa.
Our single-stage network runs much faster than most existing two-stage LiDAR detectors while surpassing their detection results, as shown in the leader board in Tab.~\ref{tab:wod_test_3d}.

\textbf{Keypoint auxiliary supervision} We devise a keypoint prediction sub-head as auxiliary supervision in the detection head inspired by~\cite{wang2020centernet3d}. We add another heatmap that predicts 4 corners and the center of every object in BEV during training. We draw the 5 keypoints of each object at the same keypoint heatmap but with a halved radius. During inference, the keypoint prediction sub-head is disabled thus does not influence the inference speed.

\subsection{Loss}
Similar to AFDet~\cite{ge2020afdet}, we apply different losses for different sub-heads. We use the Focal Loss~\cite{lin2017focal} for the heatmap prediction and keypoint auxiliary heads,  $L_1$ loss for the location offset, the $z$-axis location, the 3D object size, and orientation regression heads. We compute the target IoU by $(2 * iou - 0.5)$ $\in$ $[-1, 1]$ for the IoU-aware head, where $iou$ is the axis-aligned 3D IoU between the ground truth box and the predicted box. Smooth $L_1$ loss is used for this branch. For all sub-heads except the heatmap prediction head and keypoint auxiliary head, only $N$ foreground objects that are in the detection range are used to compute the loss. 

We use weighted sum of all losses as the final loss: 
\begin{equation}
\begin{split}
    \mathcal{L} = \mathcal{L}_{heat} + \lambda_{off}\mathcal{L}_{off} + \lambda_{z}\mathcal{L}_{z} + \lambda_{size}\mathcal{L}_{size} +\\ \lambda_{ori}\mathcal{L}_{ori} + \lambda_{iou}\mathcal{L}_{iou} + \lambda_{kps}\mathcal{L}_{kps}
\end{split}
\label{eq:loss}
\end{equation}

where $\lambda$ is the weight for each sub-head.


\section{Experiments}


\subsection{Datasets}

\subsubsection{nuScenes}
nuScenes~\cite{caesar2020nuScenes} dataset contains 700 \textit{training} sequences, 150 \textit{val} sequences and 150 \textit{test} sequences. The annotations include 10 classes. nuScenes detection score (NDS) is the main ranking metric for this dataset. We also report Mean Average Precision (mAP).

\subsubsection{Waymo Open Dataset}
The Waymo Open Dataset~\cite{sun2020scalability} contains 798 \textit{training} sequences, 202 \textit{val} sequences, and 150 \textit{test} sequences for VEHICLE, PEDESTRIAN, and CYCLIST detection. Boxes having more than five LiDAR points and not marked as LEVEL\_2 (L2) are classified as LEVEL\_1 (L1). Rest boxes that enclose at least one LiDAR point are classified as L2. All L1 and L2 boxes are considered in the L2 metric.

\subsection{Experiment Settings}
\subsubsection{nuScenes} 
We set max number of objects, max point per voxel and the max voxel number in each frame to $500$, $10$ and $160, 000$ respectively. As a convention, we accumulate 10 LiDAR sweeps to densify the point clouds~\cite{caesar2020nuScenes}. The voxel size is set to $[0.075m, 0.075m, 0.2m]$ and the training and inference ranges are set to $[\pm54m, \pm54m, (-5m, 3m)]$ for the $x,y,z$-axis. We do not use any model ensembling or Test-Time Augmentation on nuScenes Dataset.

\subsubsection{Waymo Open Dataset}
All our models only use the top LiDAR during training. The max number of objects in each frame is set to $500$. The max point per voxel and the max voxel number are set to $5$ and $250, 000$ respectively during training. We do not set a limit for the max number of points at inference. AdamW~\cite{loshchilov2018decoupled} optimizer and one-cycle policy~\cite{one_cycle} is used. The max learning rate, the division factor and momentum ranges are set to $3 \times 10^{-3}$, 10, and [0.95, 0.85]. The weight decay is fixed to 0.01. We set all $\lambda$ in Eq.~\ref{eq:loss} to 2.0. Besides, we replace max pooling with class-specific NMS for better AP. In our solution, we set the IoU threshold to 0.8, 0.55, 0.55, and $\alpha$ in the IoU rescoring branch to 0.68, 0.71, 0.65 for VEHICLE, PEDESTRIAN, and CYCLIST respectively. Models are trained with Nvidia V100 GPUs. We use the data augmentation strategy following~\cite{ge2021real}.


\begin{table}[t]
    \small
    \begin{center}
		\resizebox{0.40\textwidth}{!}{
            \begin{tabular}{ccc|cccc}
                \hline
                \tabincell{c}{Keypoint}
                & \tabincell{c}{SC-Conv}
                & \tabincell{c}{IoU}
                & ALL & VEH   & PED  & CYC  \\
                \hline
                \hline
                          &            &             & 64.65 & 64.96 & 61.82 & 67.16 \\
                \cmark    &            &             & 64.77 & 65.13 & 61.72 & 67.47 \\
                          & \cmark     &             & 65.20  & 65.24 & 62.21 & 68.14 \\
                          &            & \cmark      & 67.77 & 68.11 & 66.50 & 68.69  \\
                \cmark    & \cmark     &             & 65.58 & 65.89 & 62.56 & 68.28 \\
                \cmark    & \cmark     & \cmark      &\bf 68.77 & \bf69.22 & \bf66.95 & \bf70.12 \\
                \hline
            \end{tabular}
        }
    \end{center}
    \caption{Ablation study of the effect of the AFDetV2 improvements. ``Keypoint" represents keypoint auxiliary loss. ``SC-Conv" stands for self-calibrated convolutional backbone. All results are in 3D APH/L2 metric on full \textit{val} set.}
    \label{tbl: ablation of afdetv2 improvements}
\end{table}


\subsubsection{Different model variants}
For Waymo Open Dataset, we report 4 different variants of our model: AFDetV2-Lite, AFDetV2-Base, AFDetV2 and AFDetV2-Ens. They have the same 3D Feature Extractor, backbone and detection head. We list their detailed configurations in Tab.~\ref{tbl:model_training}.

AFDetV2-Lite for \textit{test} set evaluation is finetuned 18 epochs on \textit{trainval} set after being trained on \textit{training} set for 10 epochs. AFDetV2-Lite for \textit{val} set evaluation is 
trained 18 epochs on \textit{training} set.
For AFDetV2-Base, we first train 10 epochs on \textit{training} set and finetune 36 epochs on the whole \textit{trainval} data.
We further fintune AFDetV2-Base for another 36 epochs on the whole \textit{trainval} set using smaller grid size indicated in Tab.~\ref{tbl:model_training} to get AFDetV2.
AFDetV2-Ens is basically identical to AFDetV2 with two differences: we use all LiDAR points for AFDetV2-Ens instead of only using top-LiDAR points at inference; we use Test-Time Augmentation and model ensemble to further improve the detection accuracy for AFDetV2-Ens.

Following~\citeauthor{ge2021real}, we use Stochastic Weights Averaging (SWA)~\cite{izmailov2018averaging, zhang2020varifocalnet} in AFDetV2-Lite, AFDetV2 and AFDetV2-Ens before submitting to the official evaluation server for \textit{test} set evaluation. We never use SWA in any evaluation against \textit{val} set or in ablation studies. We don’t use SWA in AFDetV2-Base either.


\subsubsection{Model ensembling setting}
We use Test-Time Augmentation (TTA) and ensemble to improve the performance following \cite{ding20201st}. We only use yaw rotation, global scaling and translation along $z$-axis. To be specific, we use [$0\degree$, $\pm22.5\degree$, $\pm 45\degree$, $\pm 135\degree$, $\pm 157.5\degree$, $180\degree$] for yaw rotation, [0.95, 1, 1.05] for global scaling, and [-0.2m, 0m, 0.2m] for translation along $z$-axis. We merge point clouds from all LiDARs for model ensembling. The model is named as AFDetV2-Ens in Tab.~\ref{tab:wod_test_3d}. 


\subsubsection{Latency optimization} Latency optimization is allowed by the sever evaluation. Our fastest model is evaluated on the Waymo real-time 3D detection evaluation server and achieves 46.9 ms latency. We make the following efforts to accelerate our model's inference speed. First, the conversion from range images to point clouds in Cartesian coordinate and voxelization is executed on GPU. Second, except for the last layers in each sub-heads, our model is cast to half-precision. Next, we merge Batch Normalization~\cite{pmlr-v37-ioffe15} parameters into 3D Sparse Convolution and SubManifold 3D Sparse Convolution layers in 3D Feature Extractor. Finally, the keypoint auxiliary branch is disabled. Note that all other methods which were measured server latency in Tab.~\ref{tab:wod_test_3d} were optimized similarly to achieve fast inference speed (\textit{e.g.} half-precision inference).

\begin{table}[t]
    \small
    \begin{center}
		\resizebox{0.475\textwidth}{!}{
            \begin{tabular}{c|c|cccc}
                \hline
                {Model} &
                {Frames}
                & ALL & VEH   & PED  & CYC  \\
                \hline
                \hline
                $^\dagger$PointPillars~\shortcite{Lang_2019_CVPR} & 1 & - & 60.10 & 50.10 & -\\
                CenterPoint-PP~\shortcite{Yin_2021_CVPR} & 2 &60.30 & 65.50 & 55.10 & 60.20 \\
                CenterPoint-PP-2stage & 2 & 61.40 & 66.70 & 55.90 & 61.70 \\ 
                AFDetV2-PP-w/o IoU~(Ours)  & 2 & 63.15 & 67.29 & 61.84 & 60.34\\
                AFDetV2-PP~(Ours) & 2 & \bf64.57 & \bf67.43 & \bf64.02 & \bf62.26 \\
                \hline
            \end{tabular}
        }
    \end{center}
    \caption{Ablation study of PointPillars-based AFDetV2 model with and without IoU-aware rescoring. We replaced the feature extractor with a PointPillars encoder (PP) for AFDetV2. AFDetV2-PP-w/o IoU already surpasses the SOTA using PP, showing a strong baseline of classification and box regression. With IoU-aware rescoring (IoU), AFDetV2-PP shows even more significant improvement. The results are 3D APH/L2 calculated by the official Waymo evaluation metrics on the entire \textit{val} set. $\dagger$: reported by RSN~\shortcite{Sun_2021_CVPR}.}
    \label{tbl: ablation encoder}
\end{table}

\subsection{Comparison with State-of-the-art Methods}

\subsubsection{Evaluation on nuScenes \textit{test} set}
We fir compare our AFDetV2 with previous LiDAR-only non-ensemble methods on the nuScenes \textit{test} set. As shown in Tab.~\ref{tbl: nusc ablation}, our method outperforms all prior arts. Specifically, AFDetV2 surpasses HotSpotNet~\cite{chen2020object} by 2.5 NDS or 3.1 mAP. To the best of our knowledge, AFDetV2 surpasses all the published LiDAR-only non-ensemble methods on the nuScenes Detection leaderboard\textsuperscript{\rm 1}. \blfootnote{\textsuperscript{\rm 1}\url{https://www.nuscenes.org/object-detection?externalData=no&mapData=no&modalities=Lidar}, accessed on Dec 6, 2021.}


\subsubsection{Evaluation on Waymo Open Dataset \textit{val} set} We compare our AFDetV2-Lite with all published single-frame LiDAR-only non-ensemble methods on WOD \textit{val} set in Tab.~\ref{tab:wod_val_3d}. Tab.~\ref{tab:wod_val_3d} is mainly sorted by ALL APH/L2 which is specified as the official ranking metric on the dataset testing server. We can see that our AFDetV2-Lite significantly outperforms the previous state-of-the-art single-frame LiDAR-only detectors. To be specific, our AFDetV2-Lite achieves 68.77 APH/L2 for the mean of all three classes, surpassing prior art~\cite{shi2021pv} by 3.9\%.


\begin{table}[t]
    \small
    \begin{center}
		\resizebox{0.39\textwidth}{!}{
            \begin{tabular}{ccc|cccc}
                \hline
                \tabincell{c}{2S-box}&
                \tabincell{c}{2S-score}&
                \tabincell{c}{IoU}& 
                {ALL} &
                {VEH} & 
                {PED}  & 
                {CYC}  \\
                \hline
                \hline
                          &            &             & 68.81  & 69.05 & 67.39 & 70.00 \\
                \cmark &            &              & 68.79  & 68.78 & 67.55 & 70.04 \\
                \cmark & \cmark &             & 70.42 & 69.63 & 70.62 & 71.01 \\
                \cmark & \cmark & \cmark  & 71.01 & 70.10  & \bf71.71 & 71.20 \\
                 &  & \cmark  &\bf 71.10 & \bf70.73 & 71.24 & \bf71.33 \\
                \hline
            \end{tabular}
        }
    \end{center}

    \caption{The analysis of each component of the second stage and the comparison with our single-stage method. ``2S-box" and ``2S-score" mean the second stage for box refinement and score refinement, respectively. All the results are 3D APH/L2 on the entire \textit{val} set. All models take 2 frames as input. The first row is our AFDetV2 without the IoU branch. Comparing the first row and the second row, one can see that the box refinement in the second stage made a trivial change, in contrast to the score refinement. Comparing the last two rows, one can see that, for a well-designed single-stage network like AFDetV2, the box and score refinements in the second stage are no longer needed.}
    \label{tbl: ablation 2stage vs 1stage}
\end{table}

\begin{table}[t]
    \small
    \begin{center}
		\resizebox{0.46\textwidth}{!}{
            \begin{tabular}{cc|cc|cccc}
                \hline
                \tabincell{c}{2S-box}&
                \tabincell{c}{2S-score}&
                \tabincell{c}{SC-Conv} &
                \tabincell{c}{IoU}& 
                ALL &
                VEH & 
                PED  & 
                CYC \\
                
                \hline
                \hline
                    &      &            &             & 53.23 & 60.71 & 45.93 & 53.04 \\
                  \cmark  &       &            &             & 54.55 & 64.98 & 45.46 & 53.22 \\
                
                 \cmark&\cmark &            &             &58.84 & 65.62 & 48.98 & \bf61.93 \\
                \hline
                 & &       \cmark     &             &57.14  & 64.54 & 49.95 & 56.94 \\
                 \cmark& & \cmark     &             & 57.69 & \bf67.46 & 49.39 & 56.23 \\
                 \cmark &\cmark & \cmark &          & 58.78 & 65.61 & 49.02 & 61.70 \\                
                \hline

                 \cmark &\cmark & \cmark & \cmark         & 59.56 & 65.25 & 53.29 & 60.15 \\
                 & & \cmark & \cmark         & \bf59.88 & 65.36 & \bf53.64 & 60.64 \\
                \hline
            \end{tabular}
        }
    \end{center}
    \caption{Ablation study of the second-stage components for Voxel R-CNN~\shortcite{deng2021voxel} with the AFDetV2 components. All the experiments were done under the framework OpenPCDet~\shortcite{openpcdet2020}, with the default settings of Voxel R-CNN. All models were trained on 1/5 subset of the \textit{training} set and validated on 1/5 subset of \textit{val} set. All results are in 3D APH/L2. The top row shows the contributions of the box and score refinements in the second stage. It can be seen that the score refinement was the main factor of the improvement. The second row shows that by replacing the convolution blocks in the backbone with the SC-Conv block, the overall APH was improved significantly, even without the second stage. The last row shows that the second stage is unnecessary when we exploit the SC-Conv and IoU branch in the first stage.}
    \label{tbl: rethinking Voxel R-CNN}
\end{table}

\subsubsection{Evaluation on Waymo Open Dataset \textit{test} set} We also compare our AFDetV2 variants with all published methods on WOD \textit{test} set in Tab.~\ref{tab:wod_test_3d}. Tab.~\ref{tab:wod_test_3d} is split into three sub-tables. The top table is for (a) single-frame LiDAR-only non-ensemble methods; the middle table is for (b) multi-frame LiDAR-only non-ensemble methods; and the bottom table is for (c) ensemble methods. The upper row and middle row aim at onboard real-time scenario; the bottom row is for offboard use cases.

For the upper row, our AFDetV2-Lite outperforms all the previous single-frame LiDAR-only non-ensemble models with respect to both speed and accuracy. Our AFDetV2-Lite is 15.4\% faster than HIKVISION\_LiDAR~\cite{xu2021centeratt} and also does better detection than it by 2.6 ALL APH/L2.

The middle row is for multi-frame LiDAR-only non-ensemble models. As we can see our AFDetV2 surpasses all the other methods under the ALL APH/L2 metric, ranking the \nth{1} on the official Real-Time 3D Detection leaderboard\textsuperscript{\rm 2}. \blfootnote{\textsuperscript{\rm 2}\url{https://waymo.com/open/challenges/2021/real-time-3d-prediction/}, accessed on Dec 6, 2021.}

After the evaluation of real-time variants of AFDetV2 for onboard scenarios, we further explore our model for offboard use cases. Under the offboard scenario, we could leverage ample computational resources. Detection accuracy plays a much more important role than detection speed. Thus, we leverage the Test-Time Augmentation and model ensemble for our AFDetV2-Ens to achieve more accurate detection. We list all published ensemble models in the bottom row of the Tab.~\ref{tab:wod_test_3d}. Our AFDetV2-Ens outperforms all published prior arts. It also ranks the \nth{1} on the official Non-Real-Time 3D Detection leaderboard\textsuperscript{\rm 1}. \blfootnote{\textsuperscript{\rm 1}\url{https://waymo.com/open/challenges/2020/3d-detection/}, accessed on Dec 6, 2021.} While only LiDAR is used as input for AFDetV2-Ens, it still outperforms HorizonLiDAR3D which uses both camera and LiDAR input. Our AFDetV2-Ens only takes 2 frames as input. We expect that adding more frames, \textit{e.g.} 3DAL~\cite{Qi_2021_CVPR} uses $\sim${200} frames, can further improve the detection accuracy.

\begin{figure*}
\begin{center}
\includegraphics[width=0.9\textwidth]{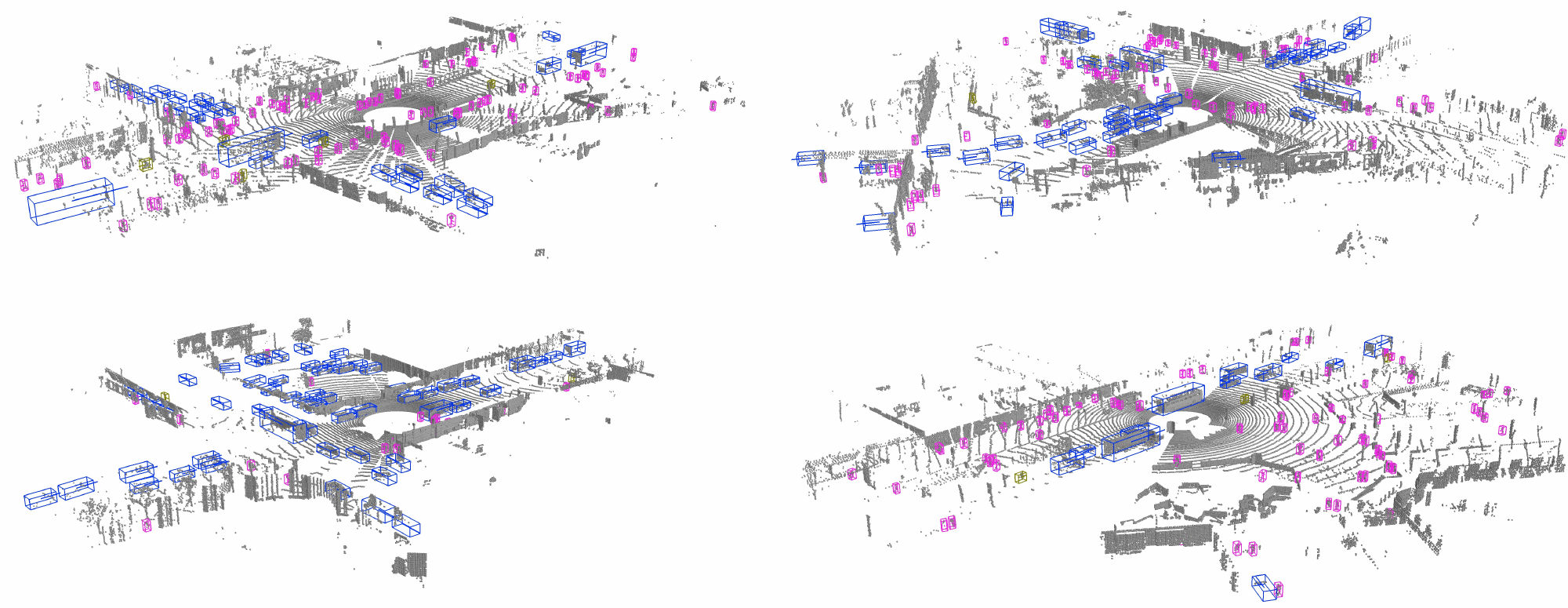}
\end{center}
\caption{The detection results of AFDetV2 on Waymo Open Dataset \textit{test} set, only bounding boxes with score larger than 0.50 are visualized. The bounding boxes of VEHICLE, PEDESTRIAN and CYCLIST are in the color blue, magenta and olive, respectively.  Additional NMS is conducted for better visualization. Better viewed in color and zoom in for more details.}
\label{fig:Final result 1}
\end{figure*}

In Fig.~\ref{fig:Final result 1}, we show some visualization results of our AFDetV2 on Waymo Open Dataset \textit{test} set.

\subsection{Waymo Real-Time 3D Detection Challenge}
The Waymo Open Dataset Real-time 3D Detection Challenge requires an algorithm to detect the 3D objects of interest as a set of 3D bounding boxes within 70 ms per frame. The model with the highest APH/L2 performance while satisfying this real-time requirement wins this challenge. In addition, the model with the lowest latency and APH/L2 $>$ 70 is given the title of ``Most Efficient Model''.

Our entry AFDetV2 won \emph{the \nth{1} place in the Real-Time 3D Detection of WOD Challenge 2021}. Our AFDetV2-Base is a faster one and is only 0.55\% lower than AFDetV2. AFDet-Base was entitled \emph{the ``Most Efficient Model'' by the WOD Challenge Sponsor}, showing a superior computational efficiency.

\subsection{Ablation Studies}
\subsubsection{Effect of AFDetV2 improvements}
To validate our method, we conduct experiments on the \textit{val} set of WOD. Three improvements are proposed in our solution which are keypoint auxiliary loss, SC-Conv backbone, and IoU rescoring. We first validate the improvement brought by each of them. From Tab.~\ref{tbl: ablation of afdetv2 improvements}, we can see that keypoint auxiliary loss brought a boost of 0.12 ALL APH/L2, utilizing SC-Conv~\cite{liu2020improving} in backbone led to an increment of 0.55 ALL APH/L2, while adding IoU rescoring enhanced the model with a considerable 3.12 ALL APH/L2. When all modules are deployed, the accuracy of the model was increased by 4.12 ALL APH/L2. We can see that the IoU rescoring brings significant improvement to the model.

To validate the generality of AFDetV2, we conduct an experiment on a different encoder. In Tab.~\ref{tbl: ablation encoder}, we replace the 3D sparse convolution feature extractor with PointPillars encoder~\cite{Lang_2019_CVPR}. Our AFDetV2 surpasses state-of-the-art PointPillars-based model~\cite{Yin_2021_CVPR} by a margin of 3.17 ALL APH/L2. Besides, the IoU-aware rescoring shows higher improvement (+1.42 ALL APH/L2) than the second stage refinement brought to the CenterPoint (+1.1 ALL APH/L2) with the same encoder.

\subsubsection{Analysis of two-stage vs. single-stage}
We conduct in-depth analysis over each component of the R-CNN style two-stage detector. In Tab.~\ref{tbl: ablation 2stage vs 1stage}, we migrate the second stage network proposed by CenterPoint~\shortcite{Yin_2021_CVPR} to AFDetV2. We feed five RoI features to the second stage for each first-stage proposal. The score and box are refined at the second stage by multi-layer perceptrons. The second stage box refinement alone would not bring performance improvement. The second stage box refinement and the second confidence score refinement together would bring 1.61 ALL APH/L2 improvement. We validate that the single-stage detector with IoU-aware rescoring can beat its corresponding two-stage detector. Furthermore, by comparing the last two rows of Tab.~\ref{tbl: ablation 2stage vs 1stage}, the single-stage detector with IoU-aware rescoring can even beat its corresponding two-stage detector with IoU-aware rescoring.
In this scenario, the second stage network is unnecessary for a well-designed single-stage network.

In Tab.~\ref{tbl: rethinking Voxel R-CNN}, we migrate our improvements (SC-Conv and IoU-aware head) to a typical state-of-the-art two-stage network, Voxel R-CNN~\shortcite{deng2021voxel}. Voxel R-CNN uses SECOND~\shortcite{yan2018second} as the first stage. Then, a voxel-RoI pooling module is applied to both of the voxel and BEV features. The acquired RoI features are fed into a second-stage network. Similar to most R-CNN style networks, the second stage has two sibling branches: one for box regression and the other for confidence prediction. The box regression branch predicts the residue from 3D region proposals to the ground truth boxes, and the confidence branch predicts the IoU-related confidence score. As shown in Tab.~\ref{tbl: rethinking Voxel R-CNN}, we divide the table into three major rows. In the top row, score and box refinement together in the second stage would bring significant improvement (+5.61 ALL APH/L2) over the baseline. In the second row, when we enhance its first stage by SC-Conv backbone, the box refinement in the second stage alone shows only minor improvement, which means bounding box regression in the first stage can achieve enough precision. In the last row, compared to the original first stage model, our enhanced single-stage model with SC-Conv and IoU rescoring shows higher improvement (+6.65 ALL APH/L2) than adding the second stage (+5.61 ALL APH/L2). Also, an extra second stage is useless to the overall performance, which is similar to our findings in Tab.~\ref{tbl: ablation 2stage vs 1stage}. Once again, we prove that the second stage is unnecessary if we strengthen and fully utilize the first stage by our proposed methods.

\section{Conclusion}
We have proposed a real-time single-stage anchor-free 3D object detection model named AFDetV2. We have conducted extensive experiments to show that the second stage is unnecessary if we strengthen and fully utilize the first stage. Our model achieves the state-of-the-art performance on Waymo Open Dataset and nuScenes Dataset with high inference speed.

\bibliography{aaai22}
\end{document}


\maketitle

\newcommand\blfootnote[1]{%
  \begingroup
  \renewcommand\thefootnote{}\footnote{#1}%
  \addtocounter{footnote}{-1}%
  \endgroup
}
\appendix


\section{Overview}
In this document, we provide more results and details of our method. To be specific, Sec.~\ref{more_wod_results} presents more evaluation results on Waymo Open Dataset~\cite{sun2020scalability}. We show some visualization results in Sec.~\ref{visualization}. More details of 3D Feature Extractor, different model variants and data augmentation are shown in Sec.~\ref{feature_extractor}, Sec.~\ref{model_variants} and Sec.~\ref{data_aug}, respectively.

\section{More Evaluation Results}
\label{more_wod_results}
In Tab.~\ref{tab:wod_val_3d} we include the comparison for PEDESTRIAN and CYCLIST on the Waymo Open Dataset \textit{val} set. Our AFDetV2-Lite outperforms CenterPoint~\cite{Yin_2021_CVPR} on PEDESTRIAN detection by 1.65 APH/L2 and outperforms PV-RCNN-v2~\cite{shi2021pv} on CYCLIST detection by a large margin of 4.95 APH/L2.

In Tab.~\ref{tab:wod_test_3d} we also include the comparison for PEDESTRIAN and CYCLIST on the Waymo Open Dataset \textit{test} set. As we can see our models consistently surpass all prior arts under different settings for PEDESTRIAN and CYCLIST classes. While only LiDAR is used as input for our AFDetV2-Ens, it still outperforms HorizonLiDAR3D which uses both camera and LiDAR input on PEDESTRIAN and CYCLIST detection.

\section{Qualitative Results}
\label{visualization}
In Fig.~\ref{fig:Final result 1}, we show some visualization results of our AFDetV2 variant on Waymo Open Dataset \textit{test} set. The bounding boxes of VEHICLE, PEDESTRIAN and CYCLIST are in the color blue, magenta and olive, respectively.

\begin{figure}[t]
\centering
\begin{subfigure}[b]{.45\columnwidth}
\centering
\includegraphics[width=.7\linewidth]{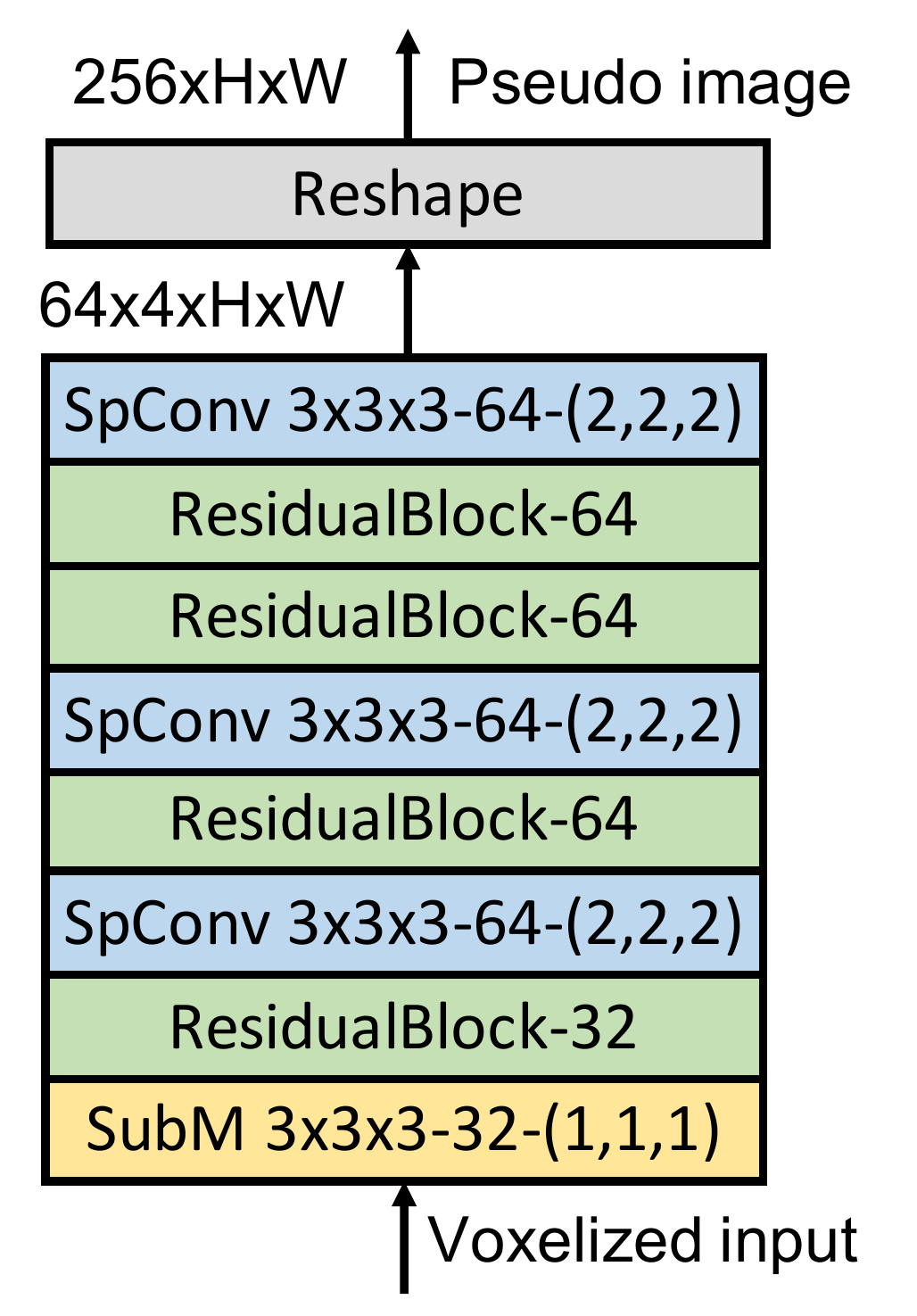}
\caption{3D Feature Extractor}\label{fig:feature_extractor}
\end{subfigure}
\hfill
\begin{subfigure}[b]{.45\columnwidth}
\centering
\includegraphics[width=.7\linewidth]{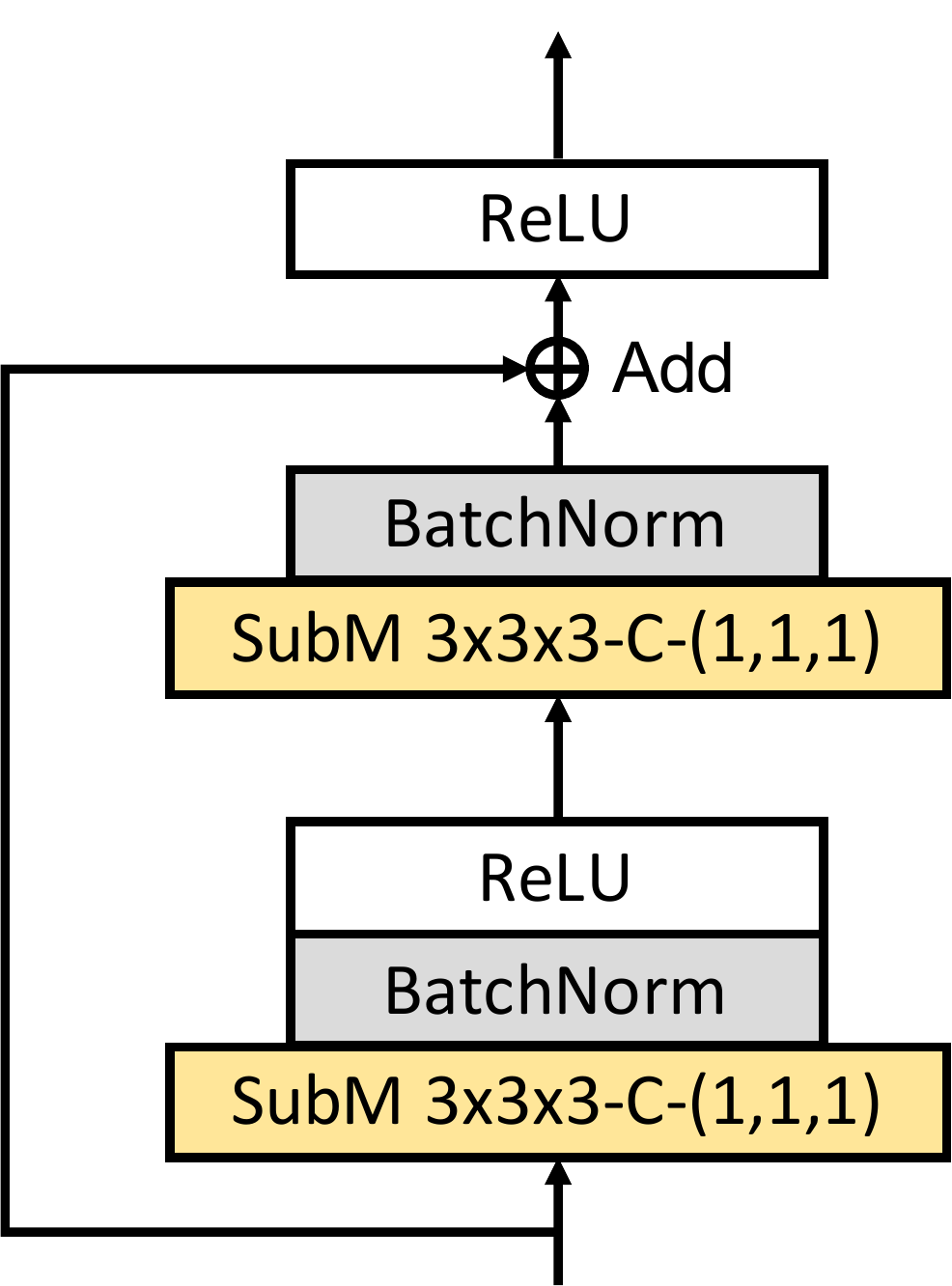}
\caption{Residual block}\label{fig:residual_block}
\end{subfigure}
\caption{(a) The structure of the 3D Feature Extractor. ``SubM" stands for sub-manifold sparse convolutional layer~\cite{graham2017submanifold} and ``SpConv" stands for 3D sparse convolutional layer~\cite{yan2018second}. The format of the layer setting follows ``kernel size-channels-(strides)", \ie $k_W \times k_H \times k_D$-$C$-($s_H$,$s_W$,$s_D$). For residual blocks, only channels are shown. After the backbone, channel dimension and $D$ dimension are reshaped together to form a pseudo image. (b) The structure of a residual block with $C$ channels.}
\label{fig:backbone}
\end{figure}


\section{More Details of 3D Feature Extractor}
\label{feature_extractor}
Our 3D Feature Extractor uses 3D sparse convolutional layers~\cite{yan2018second} and submanifold sparse convolutional layers~\cite{graham2017submanifold}. The 3D sparse convolutional layer is used to downsample the 3D feature maps with a stride of 2 while submanifold sparse convolutional layer is used to extract features at different strides by composing residual blocks. The whole feature extractor has a total stride of 8 for all three dimensions. Fig.~\ref{fig:backbone} shows the detailed structure of the proposed 3D Feature Extractor. After 3D feature extraction, the $z$-axis dimension and the channel dimension are reshaped together to reduce the dimension of the final feature map. In the end, a Bird's Eye View (BEV) pseudo image is generated and is sent into the self-calibrated convolutional backbone.

\begin{table*}[h]
\begin{center}

\vspace{5pt}
\setlength\tabcolsep{2pt}
\begin{tabular}{l|c|c|c c c|c c c|c c c|c c c}
\hline
\multirow{2}{*}{Names} &\multirow{2}{*}{Sensors} &\multirow{2}{*}{Frames} &\multirow{2}{*}{Ens} & Usage  &Server &\multicolumn{3}{c|}{Training} &\multicolumn{3}{c|}{Inference}  &\multicolumn{3}{c}{Grid} \\
&&&&Scenario &Latency &\multicolumn{3}{c|}{Range} &\multicolumn{3}{c|}{Range} &\multicolumn{3}{c}{Size}\\
\hline
\hline
AFDetV2-Lite  & LT &1  &\xmark &Onboard  &46.90     &75.2  &75.2  &-2  &75.2  &75.2  &4 &0.1  &0.1  &0.15  \\
AFDetV2-Base  & LT &2  &\xmark &Onboard  &55.86     &75.2  &75.2  &-2  &75.2  &75.2  &4 &0.1  &0.1  &0.15  \\
AFDetV2       & LT &2  &\xmark &Onboard  &60.06     &75.2  &73.6  &-2  &80.0  &76.16 &4 &0.1  &0.08 &0.15  \\
AFDetV2-Ens   & L  &2  &\cmark &Offboard &-   &75.2  &73.6  &-2  &80.0  &76.16     &4     &0.1       &0.08    &0.15  \\

\hline
\end{tabular}
\end{center}
\caption{The configurations of different variants of our model. ``L" and ``LT" mean ``all LiDARs" and ``top-LiDAR only", separately. ``Ens" is short for ensemble. ``Server Latency" is the latency measured by the official testing server in milliseconds. Values in training range, inference range and grid size columns are in meters with respect to $x, y, z$-axes, respectively. Our AFDetV2-Ens uses Test-Time Augmentation and model ensemble for better detection accuracy and is suitable for offboard usage (\textit{e.g.} auto labeling). The other 3 variants are real-time and suitable for onboard usage.}
\label{tbl:model_training}
\end{table*}

\begin{table*}
    \definecolor{Gray}{gray}{0.9}
    \newcolumntype{g}{>{\columncolor{Gray}}c}
	\begin{center}
		\resizebox{\textwidth}{!}{
			\begin{tabular}{c|c|cg|cc|cc|cc}
				\hline
				\multirow{2}{*}{Methods} & 
				\multirow{2}{*}{\# Stages} & 			
				\multicolumn{2}{c|}{ALL (3D AP/APH)} & \multicolumn{2}{c|}{VEH (3D AP/APH)} & 
				\multicolumn{2}{c|}{PED (3D AP/APH)} & \multicolumn{2}{c}{CYC (3D AP/APH)} \\
				&& L1 & L2 & L1 & L2 & L1 & L2 & L1 & L2 \\
				\hline
				\hline
				StarNet~\shortcite{ngiam2019starnet}        & Two   & - & - & 53.70/- & - & 66.80/- & - & - & -  \\ 
				PPBA~\shortcite{cheng2020improving}         & Single & - & - &62.40/- & - & 66.00/- & - & -  & -\\
				MVF~\shortcite{zhou_2020_mvf}               & Single & - & - & 62.93/- & - & 65.33/- & - & - & -\\
				CVCNet~\shortcite{chen2020every}            & Single & - & - & 65.20/- & - & - & - & - & - \\
				3D-MAN~\shortcite{yang20213d}               & Two      & - & - & 69.03/68.52 & 60.16/59.71 & 71.71/67.74 & 62.58/59.04 & - & - \\
				RCD~\shortcite{bewley2020range}             & Two    & - & - & 69.59/69.16 & - & - & - & - & - \\
				Pillar-based~\shortcite{wang2020pillar}     & Single & - & - & 69.80/- & - & 72.51/- & - & - & - \\
				$^\dagger$SECOND~\shortcite{yan2018second}          & Single      & 67.20/63.05 & 60.97/57.23 & 72.27/71.69 & 63.85/63.33 & 68.70/58.18 & 60.72/51.31 & 60.62/59.28 & 58.34/57.05 \\
				$^\ddagger$PointPillars~\shortcite{Lang_2019_CVPR}  & Single      &  68.87/63.33 & 62.63/57.53 & 71.60/71.00 & 63.10/62.50 & 70.60/56.70 & 62.90/50.20 & 64.40/62.30 & 61.90/59.90 \\
				LiDAR R-CNN~\shortcite{li2021lidar}         & Two    & 71.10/66.20 & 64.63/60.10  & 73.50/73.00 & 64.70/64.20 & 71.20/58.70 & 63.10/51.70 & 68.60/66.90 & 66.10/64.40 \\
				RangeDet~\shortcite{fan2021rangedet}        & Single & 71.53/- & - &72.85/ & - & 75.94/- & - & 65.80/- & - \\
				MVF++~\shortcite{Qi_2021_CVPR}              & Single & - & - & 74.64/- & - & 78.01/- & - & - & - \\
				RSN~\shortcite{Sun_2021_CVPR}               & Two    & - & - & 75.10/74.60 & 66.00/65.50 & 77.80/72.70 & 68.30/63.70 & - & - \\
				$\text{H}^2$3D R-CNN~\shortcite{deng2021multi}   & Two & - & - & 75.15/- & 66.14/- & - & - & - & - \\
				M3DETR~\shortcite{guan2021m3detr}           & Two    & - & - & 75.71/75.08 & 66.58/66.02 & - & - & - & - \\
				Voxel R-CNN~\shortcite{deng2021voxel}       & Two    & - & - & 75.59/- & 66.59/- & - & - & - & - \\
				CenterPoint~\shortcite{Yin_2021_CVPR}       & Two    & - & - & 76.70/76.20 & 68.80/68.30 & 79.00/72.90 & 71.00/65.30 & - & - \\
				PV-RCNN~\shortcite{shi2020pv}               & Two    & 73.44/69.63 & 66.80/63.33 & 77.51/76.89 & 68.98/68.41 & 75.01/65.65 & 66.04/57.61 & 67.81/66.35 & 65.39/63.98 \\
				Part-$A^2$~\shortcite{shi2020part}         & Two     & 73.63/70.25 & 66.93/63.84 & 77.05/76.51 & 68.47/67.97 & 75.24/66.87 & 66.18/58.62 & 68.60/67.36 & 66.13/64.93 \\
				PV-RCNN-v2~\shortcite{shi2021pv}            & Two    & 74.81/71.00 & 68.42/64.87 & \textbf{78.79/78.21} & \textbf{70.26/69.71} & 76.67/67.15 & 68.51/59.72 & 68.98/67.63 & 66.48/65.17 \\
				AFDetV2-Lite (Ours)                              & Single & \textbf{77.18/74.83} & \textbf{70.97/68.77} & 77.64/77.14 & 69.68/69.22 & \textbf{80.19/74.62} & \textbf{72.16/66.95} & \textbf{73.72/72.74} & \textbf{71.06/70.12} \\
				\hline
			\end{tabular}
		}
	\end{center}
	\caption{The single-frame LiDAR-only non-ensemble 3D AP/APH performance comparison on the Waymo Open Dataset \textit{val} set. ``ALL" stands for the mean of all three classes. The table is mainly sorted by ALL APH/L2 which is the official ranking metric. $\dagger$: re-implemented by PV-RCNN++~\shortcite{shi2021pv}. $\ddagger$: re-implemented by LiDAR R-CNN~\shortcite{li2021lidar}.}
	\label{tab:wod_val_3d}
\end{table*} 

\begin{table*}
    \definecolor{Gray}{gray}{0.9}
    \newcolumntype{g}{>{\columncolor{Gray}}c}
	\begin{center}
		\resizebox{\textwidth}{!}{
			\begin{tabular}{c|c|c|c|c|cg|cc|cc|cc}
				\hline
				\multirow{2}{*}{Methods} & 
				\multirow{2}{*}{Sensors} &
				\multirow{2}{*}{Frames} &
				\multirow{2}{*}{Ens} &
				Server & 			
				\multicolumn{2}{c|}{ALL (3D AP/APH)} & \multicolumn{2}{c|}{VEH (3D AP/APH)} & 
				\multicolumn{2}{c|}{PED (3D AP/APH)} & \multicolumn{2}{c}{CYC (3D AP/APH)} \\
				&&&& Latency & L1 & L2 & L1 & L2 & L1 & L2 & L1 & L2 \\
				\hline
				\hline 
				StarNet~\shortcite{ngiam2019starnet} & - & 1 & \xmark & - & - & - & 61.50/61.00 & 54.90/54.50 & 67.80/59.90 & 61.10/54.00 & - & -  \\ 
				PPBA~\shortcite{cheng2020improving} & - & 1 & \xmark & - & - & - & 67.50/67.00 & 59.60/59.10 & 69.70/61.70 & 63.00/55.80 & - & - \\
				$^\dagger$PointPillars~\shortcite{Lang_2019_CVPR} & LT & 1 & \xmark & - & - & - & 68.60/68.10 & 60.50/60.10 & 68.00/55.50 & 61.40/50.10 & - & - \\ 
				RCD~\shortcite{bewley2020range} & - & 1 & \xmark & - & - & - & 71.97/71.59 & 65.06/64.70 & - & - & - & - \\
				Pseudo-Labeling~\shortcite{caine2021pseudo}   & - & 1 & \xmark &- & - & - & 74.00/73.60 & - & 69.80/57.90 & - & - & - \\
				M3DETR~\shortcite{guan2021m3detr} & - & 1 & \xmark & - & 71.05/67.09 & 65.50/61.92 & 77.75/77.17 & 70.63/70.06 & 68.10/58.87 & 60.57/52.37 & 67.28/65.69 & 65.31/63.75 \\ 
				Light-FMFNet~\shortcite{Murhij_2021_challenge} & L & 1 & \xmark & 62.31 & 71.24/67.26 & 65.88/62.18 & 77.85/77.30 & 70.16/69.65 & 69.52/59.78 & 63.62/54.61 & 66.34/64.69 & 63.87/62.28 \\
				HIKVISION\_LiDAR~\shortcite{xu2021centeratt} & L & 1 & \xmark & 54.13 & 75.19/72.58 & 69.74/67.29 & 78.63/78.14 & 71.06/70.60 & 76.00/69.90 & 69.82/64.11 & 70.94/69.70 & 68.35/67.15 \\
				CenterPoint~\shortcite{Yin_2021_CVPR} & - & 1 & \xmark & - & - & - & 80.20/79.70 & 72.20/71.80 & 78.30/72.10 & 72.20/66.40 & - & - \\
				AFDetV2-Lite~(Ours) & LT & 1 & \xmark & \textbf{46.90} & \textbf{77.56/75.20} & \textbf{72.18/69.95} & \textbf{80.49/80.03} & \textbf{72.98/72.55} & \textbf{79.76/74.35} & \textbf{73.71/68.61} & \textbf{72.43/71.23} & \textbf{69.84/68.67} \\
				\hline
				3D-MAN~\shortcite{yang20213d} & L & 15 & \xmark & - & - & - & 78.71/78.28 & 70.37/69.98 & 69.97/65.98 & 63.98/60.26 & - & - \\
				RSN~\shortcite{Sun_2021_CVPR} & LT & 3 & \xmark & - & - & - & 80.70/80.30 & 71.90/71.60 & 78.90/75.60 & 70.70/67.80 & - & - \\ 
				X\_Autonomous3D~\shortcite{Liu_2021_challenge} & L & 2 & \xmark & 68.42 & 77.54/75.61 & 72.29/70.46 & 81.49/81.02 & 74.04/73.60 & 78.17/73.93 & 72.29/68.27 & 72.96/71.88 & 70.55/69.50 \\
				CenterPoint~\shortcite{Yin_2021_CVPR} & L & 2 & \xmark & - & 78.71/77.18 & 73.38/71.93 & 81.05/80.59 & 73.42/72.99 & 80.47/77.28 & 74.56/71.52 & 74.60/73.68 & 72.17/71.28 \\
				AFDetV2-Base~(Ours) & LT & 2 & \xmark & \textbf{55.86} & 79.24/77.67 & 74.06/72.57 & 81.27/80.82 & 73.89/73.46 & 81.08/77.87 & 75.34/72.29 & 75.35/74.33 & 72.96/71.97 \\
				CenterPoint++~\shortcite{Yin_2021_challenge} & LT & 3 & \xmark & 57.12 & 79.41/77.96 & 74.22/72.82 & \textbf{82.78/82.33} & \textbf{75.47/75.05} & 81.07/\textbf{78.21} & 75.13/72.41 & 74.40/73.33 & 72.04/71.01 \\
				AFDetV2~(Ours) & LT & 2 & \xmark & 60.06 & \textbf{79.77/78.21} & \textbf{74.60/73.12} & 81.65/81.22 & 74.30/73.89 & \textbf{81.26}/78.05 & \textbf{75.47/72.41} & \textbf{76.41/75.37} & \textbf{74.05/73.04} \\
				\hline
				TS-LidarDet~\shortcite{Wang_2020_challenge} & L & 1 & \cmark & - & 74.87/71.05 & 69.10/65.53 & 80.75/80.18 & 72.65/72.12 & 74.45/65.01 & 68.10/59.32 & 69.42/67.97 & 66.55/65.16 \\
				RSN Ens~\shortcite{Sun_2021_CVPR} & LT & 3 & \cmark & - & - & - & 81.38/80.97 & 72.80/72.43 & 82.41/77.98 & 74.75/70.68 & - & - \\
				PV-RCNN Ens~\shortcite{shi_2020_challenge} & L & 2 & \cmark & - & 78.82/76.90 & 73.35/71.52 & 81.06/80.57 & 73.69/73.23 & 80.31/76.28 & 73.98/70.16 & 75.10/73.84 & 72.38/71.16 \\
				3DAL~\shortcite{Qi_2021_CVPR} & L & $\sim${200}  & \cmark & - & - & - & \textbf{85.84/85.46} & 77.24/76.91 & - & - & - & - \\ 
				HorizonLiDAR3D~\shortcite{ding20201st} & CL & 5 & \cmark & - & 83.28/81.85 & 78.49/77.11 & 85.09/84.68 & 78.23/77.83 & 85.03/82.10 & 79.32/76.50 & 79.73/78.78 & 77.91/76.98 \\
				AFDetV2-Ens~(Ours) & L & 2 & \cmark & - & \textbf{84.07/82.63} & \textbf{79.04/77.64}  & 85.80/85.41  & \textbf{78.71/78.34} & \textbf{85.22/82.16} & \textbf{79.71/76.75} & \textbf{81.20/80.30} & \textbf{78.70/77.83} \\
				\hline
			\end{tabular}
		}
	\end{center}
	\caption{The 3D AP/APH performance comparison on the Waymo Open Dataset \textit{test} set. ``L", ``LT" and ``CL" mean ``all LiDARs", ``top-LiDAR only" and ``camera and all LiDARs", separately. ``ALL" stands for the mean of all three classes. The table is mainly sorted by ALL APH/L2 which is the official ranking metric. ``Ens" is short for ensemble.  ``Server Latency" is the latency measured by the official testing server in milliseconds. The table is split into three major rows. The upper row is for single-frame LiDAR-only non-ensemble models; the middle row is for multi-frame LiDAR-only non-ensemble models; the bottom row is for ensembled models. Our models consistently outperform previous state-of-the-art methods under different settings. $\dagger$: re-implemented by RSN~\shortcite{Sun_2021_CVPR}.}
	\label{tab:wod_test_3d}
\end{table*} 

\begin{figure*}
\begin{center}
\includegraphics[width=1\textwidth]{LaTeX/images/merged_4_image.pdf}
\end{center}
\caption{The detection results of AFDetV2 on Waymo Open Dataset \textit{test} set, only bounding boxes with score larger than 0.50 are visualized. The bounding boxes of VEHICLE, PEDESTRIAN and CYCLIST are in the color blue, magenta and olive, respectively.  Additional NMS is conducted for better visualization. Better viewed in color and zoom in for more details.}
\label{fig:Final result 1}
\end{figure*}

\section{More Details of Different Model Variants}
\label{model_variants}
In the paper, we report the performance of 4 different variants of our model: AFDetV2-Lite, AFDetV2-Base, AFDetV2 and AFDetV2-Ens. They have the same 3D Feature Extractor, backbone and detection head. We list their detailed configurations in Tab.~\ref{tbl:model_training}.

AFDetV2-Lite for \textit{test} set evaluation is finetuned 18 epochs on \textit{trainval} set after being trained on \textit{training} set for 10 epochs. AFDetV2-Lite for \textit{val} set evaluation is 
trained 18 epochs on \textit{training} set.
For AFDetV2-Base, we first train 10 epochs on \textit{training} set and finetune 36 epochs on the whole \textit{trainval} data.
We further fintune AFDetV2-Base for another 36 epochs on the whole \textit{trainval} set using smaller grid size indicated in Tab.~\ref{tbl:model_training} to get AFDetV2.
AFDetV2-Ens is basically identical to AFDetV2 with two differences: we use all LiDAR points for AFDetV2-Ens instead of only using top-LiDAR points at inference; we use Test-Time Augmentation and model ensemble to further improve the detection accuracy for AFDetV2-Ens.

We use Stochastic Weights Averaging (SWA)~\cite{izmailov2018averaging, zhang2020varifocalnet} in AFDetV2-Lite, AFDetV2 and AFDetV2-Ens before submitting to the official evaluation server for \textit{test} set evaluation. Specifically, we train 5 extra epochs with 1/10 of the original learning rate. In each epoch, the learning rate first warms up from $3 \times 10^{-5}$ to  $3 \times 10^{-4}$ and then decreases at each iteration from  $3 \times 10^{-4}$ to $3 \times 10^{-9}$. We calculate the average of 5 checkpoints and finetune another 1 epoch to fix the Batch Normalization~\cite{pmlr-v37-ioffe15} parameters. We never use SWA in any evaluation against \textit{val} set or in ablation studies. We don’t use SWA in AFDetV2-Base either.

\section{More Details of Data Augmentation}
\label{data_aug}
We use data augmentation strategy following~\cite{yan2018second,zhu2019class,ding20201st}. We first generate a database containing labels and the associated point cloud data. 15, 10 and 10 ground truth samples for VEHICLE, PEDESTRIAN and CYCLIST are randomly sampled and placed into the current frame during training. Besides, we do random flipping along $x,y$-axes, global rotation with a uniform distribution of $\mathcal{U}\left ( -\frac{\pi}{4}, \frac{\pi}{4} \right )$, global scaling of $\mathcal{U}\left ( 0.95, 1.05 \right )$ and global translation along $x,y,z$-axes of $\mathcal{U}\left ( -0.2m, 0.2m \right )$. Third, random rotation noise with a uniform distribution of  $\mathcal{U}\left ( -\frac{\pi}{20}, \frac{\pi}{20} \right )$ and random location noise with a Gaussian distribution $\mathcal{N}\left ( 0.0, ~0.1 \right )$ are applied to each instance.

\bibliography{aaai22}